\RequirePackage{fix-cm}
\documentclass[review,twocolumn]{svjour3}          
\smartqed  

\usepackage{amsmath,amsfonts,mathtools,caption}

\newcommand{\rmvp}{RMVP3D~\citep{han2024nersp}\xspace}
\newcommand{\smvp}{SMVP3D~\citep{han2024nersp}\xspace}

\newcommand{\pandora}{PANDORA~\citep{dave2022pandora}\xspace}
\newcommand{\mvas}{MVAS~\citep{cao2023multi}\xspace}
\newcommand{\nero}{NeRO~\citep{liu2023nero}\xspace}

\newcommand{\nersp}{NeRSP~\citep{han2024nersp}\xspace}

\newcommand{\pisr}{PISR~\citep{chen2024pisr}\xspace}
\newcommand{\dr}{3DGS-DR~\citep{ye20243d}\xspace}
\newcommand{\relight}{Re-3DGS~\citep{gao2023relightable}\xspace}
\newcommand{\gs}{Gaussian Surfels~\citep{dai2024high}\xspace}
\newcommand{\polgs}{PolGS++\xspace}

\newcommand{\gsror}{GS-ROR~\citep{zhu2025gs}\xspace}
\newcommand{\V}[1]{\ensuremath{\mathbf{#1}}} \usepackage{pifont}
\usepackage{multirow}
\usepackage{overpic}
\usepackage{epsfig}
\usepackage{times}
\usepackage{url}            
\usepackage{booktabs}       
\usepackage{amsfonts}       
\usepackage{nicefrac}       
\usepackage{microtype}      
\usepackage{xcolor}         
\usepackage{xspace}
\usepackage{graphicx}
\usepackage{wrapfig}
\usepackage{amsmath}
\usepackage{amssymb}
\usepackage{bm}
\usepackage{tabularx}
\usepackage{paralist}
\usepackage{natbib}
\usepackage{bbding}
\usepackage[linesnumbered,ruled,vlined]{algorithm2e}
\definecolor{cvprblue}{rgb}{0.21,0.49,0.74}
\usepackage[pagebackref,breaklinks,colorlinks,citecolor=cvprblue]{hyperref}

\usepackage{titlesec}
\titleformat{\paragraph}[runin]{\normalfont\normalsize\bfseries}{\theparagraph}{}{}[\quad]

\newcommand{\Tref}[1]{Table~\ref{#1}}
\newcommand{\Eref}[1]{Equation~(\ref{#1})}
\newcommand{\Fref}[1]{Figure~\ref{#1}}
\newcommand{\Sref}[1]{Section~\ref{#1}}

\newcommand{\eref}[1]{Eq.~(\ref{#1})}
\newcommand{\fref}[1]{Fig.~\ref{#1}}
\newcommand{\sref}[1]{Sec.~\ref{#1}}

\makeatletter
\@ifundefined{blfootnote}{%
\newcommand{\blfootnote}[1]{%
	\begingroup
	\renewcommand\thefootnote{}%
	\footnote{\noindent\setlength{\parindent}{0pt}\ignorespaces#1}%
	\addtocounter{footnote}{-1}%
	\endgroup
}%
}{%
\renewcommand{\blfootnote}[1]{%
	\begingroup
	\renewcommand\thefootnote{}%
	\footnote{\noindent\setlength{\parindent}{0pt}\ignorespaces#1}%
	\addtocounter{footnote}{-1}%
	\endgroup
}%
}
\makeatother

\makeatletter
\DeclareRobustCommand\onedot{\futurelet\@let@token\@onedot}
\def\@onedot{\ifx\@let@token.\else.\null\fi\xspace}
 
\def\ie{\emph{i.e}\onedot}

\makeatother
\begin{document}
\sloppy
\title{PolGS++: Physically-Guided Polarimetric Gaussian Splatting for Fast Reflective Surface Reconstruction}


\author{Yufei Han$^1$, Chu Zhou$^2$, Youwei Lyu$^3$,  Qi Chen$^1$, Si Li$^1$,
Boxin Shi$^{4,5}$, Yunpeng Jia$^1$, Heng Guo$^{1*}$, Zhanyu Ma$^1$}

\authorrunning{Yufei Han et al.} 

\date{Received: date / Accepted: date}

\twocolumn[{\renewcommand\twocolumn[1][]{#1}
    \maketitle
    \centering
    \vspace{-0.5em}
    \begin{minipage}[b]{\textwidth}
       \begin{overpic}[width=\textwidth]{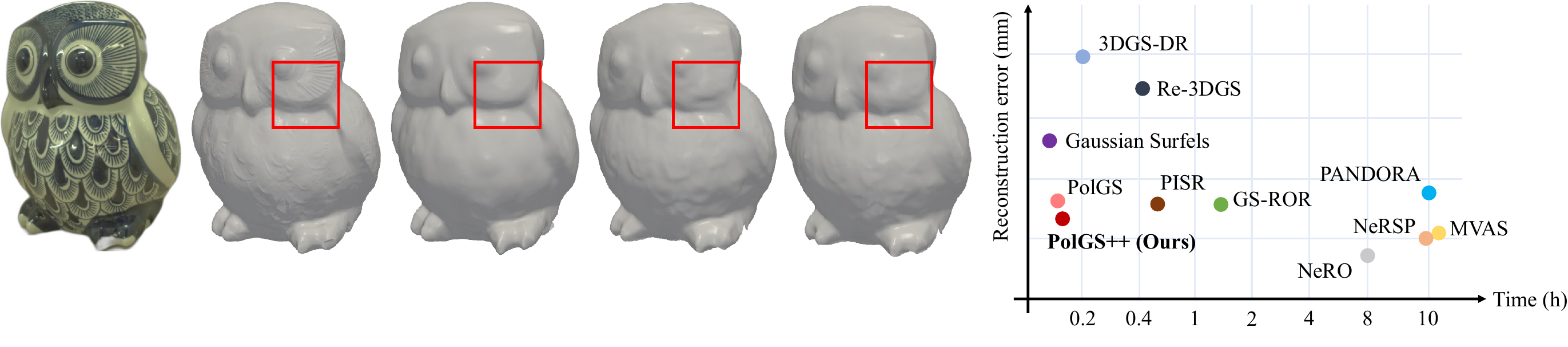}
        \put(3.6,3.5){\color{black}{\fontsize{8.7pt}{1pt}\selectfont Input}}
        \put(16.2,3.9){\color{black}{\fontsize{8.7pt}{1pt}\selectfont NeRO}}
        \put(26.5,3.9){\color{black}{\fontsize{8.7pt}{1pt}\selectfont PANDORA}}
        \put(41.5,3.9){\color{black}{\fontsize{8.7pt}{1pt}\selectfont PolGS}}
        \put(52.5,3.9){\color{black}{\fontsize{8.7pt}{1pt}\selectfont PolGS++}}

        \put(13,2.3){\color{black}{\fontsize{8pt}{1pt}\selectfont \cite{liu2023nero}}}
        \put(25,2.3){\color{black}{\fontsize{8pt}{1pt}\selectfont \cite{dave2022pandora}}}
        \put(38.5,2.3){\color{black}{\fontsize{8pt}{1pt}\selectfont \cite{han2025polgs}}}
        \put(54,2.3){\color{black}{\fontsize{8pt}{1pt}\selectfont Ours}}

        \put(2.35,1.3){\color{black}{\fontsize{8.7pt}{1pt}\selectfont $35$ views}}
        \put(15.5,0.5){\color{black}{\fontsize{8.7pt}{1pt}\selectfont $8$ hours}}
        \put(27.5,0.5){\color{black}{\fontsize{8.7pt}{1pt}\selectfont $10$ hours}}
        \put(41.5,0.5){\color{black}{\fontsize{8.7pt}{1pt}\selectfont $7$ min}}
        \put(53,0.5){\color{black}{\fontsize{8.7pt}{1pt}\selectfont $10$ min}}

        \end{overpic}
    \end{minipage}

    \vspace{-0.7 em}
    \captionof{figure}{
		Comparison of efficiency and accuracy on reflective surface reconstruction. 
		Our method takes only 10 minutes while achieving a shape reconstruction accuracy~(measured by Chamfer Distance in millimeters) comparable to that of the 
		neural implicit surface representation-based method~\nero.}
    \vspace{0.4 em}
    \label{fig.teaser}

}]
 
\begin{abstract}
\blfootnote{\vspace{-1.5em}\\
Corresponding author: Heng Guo\\
\email{guoheng@bupt.edu.cn} \vspace{0.1cm}\\ 
$1$ \, Beijing University of Posts and Telecommunications, China\\
$2$ \, National Institute of Informatics, Japan\\
$3$ \, vivo BlueImage Lab, vivo Mobile Communication Co., Ltd., Shanghai, China\\
$4$ \, State Key Laboratory for Multimedia Information Processing, School of Computer Science, Peking University, China\\
$5$ \, National Engineering Research Center of Visual Technology, School of Computer Science, Peking University, China
}
	Accurate reconstruction of reflective surfaces remains a fundamental challenge in computer vision, with broad applications in real-time virtual reality and digital content creation. Although 3D Gaussian Splatting (3DGS) enables efficient novel-view rendering with explicit representations, its performance on reflective surfaces still lags behind implicit neural methods, especially in recovering fine geometry and surface normals. To address this gap, we propose \polgs, a physically-guided polarimetric Gaussian Splatting framework for fast reflective surface reconstruction. Specifically, we integrate a polarized BRDF (pBRDF) model into 3DGS to explicitly decouple diffuse and specular components, providing physically grounded reflectance modeling and stronger geometric cues for reflective surface recovery. 
	Furthermore, we introduce a \emph{depth-guided visibility mask acquisition} mechanism that enables angle-of-polarization (AoP)-based tangent-space consistency constraints in Gaussian Splatting without costly ray-tracing intersections. 
	This physically guided design improves reconstruction quality and efficiency, requiring only 	about 10 minutes of training. Extensive experiments on both synthetic and real-world datasets validate the 	effectiveness of our method.
Code page: \href{https://github.com/PRIS-CV/PolGS\_plus}{https://github.com/PRIS-CV/PolGS\_plus}.
\end{abstract}
\vspace{-2em}
 \section{Introduction}

Accurate and robust reconstruction of reflective surfaces is essential for applications such as real-time 
virtual reality, digital content creation, and inverse rendering. Reflective materials remain challenging for 
3D reconstruction because their view-dependent specular appearance violates the photometric consistency 
assumptions widely used by multi-view stereo and neural rendering methods. Therefore, jointly modeling 
light--surface interaction and geometric consistency with physically grounded constraints remain a fundamental open problem.

Most existing reflective-surface reconstruction methods are built on implicit neural representations~\citep{mildenhall2020nerf}, 
such as Ref-NeRF~\citep{verbin2022ref} and NeRO~\citep{liu2023nero}. 
These methods model view-dependent reflectance and typically recover geometry through an implicit Signed Distance Field (SDF). 
Polarization-based approaches further improve reconstruction quality by injecting physically grounded cues. 
Some methods~\citep{dave2022pandora,han2024nersp,li2024neisf,li2025neisf++} leverage the polarimetric bidirectional reflectance distribution function (pBRDF)~\citep{baek2018simultaneous}, 
while others~\citep{zhao2022polarimetric,cao2023multi,han2024nersp,chen2024pisr} exploit angle-of-polarization (AoP) observations to constrain surface normals and reduce geometric ambiguity. 
Nevertheless, these methods still rely heavily on implicit optimization with volumetric sampling and MLP-based fitting, leading to high computational overhead and slow reconstruction.

In contrast, 3D Gaussian Splatting (3DGS)~\citep{kerbl20233d} provides an explicit representation with highly efficient rendering. 
However, this efficiency does not directly translate to strong shape modeling, since 3DGS lacks direct supervision and structural regularization for surface normals. 
Consequently, geometric constraints remain weak and reconstructions are often unreliable; this limitation is more pronounced on reflective surfaces.
Therefore, most 3DGS-based reconstruction methods mainly focus on diffuse scenes~\citep{chen2023neusg,yu2024gsdf,lyu20243dgsr,guedon2024sugar,dai2024high,huang20242d}. 

To address these issues, we propose \textbf{\polgs}, a physically-guided polarimetric Gaussian Splatting framework for fast reflective surface reconstruction. 
By leveraging \emph{photometric} cues from polarization, we build a pBRDF model in explicit 3DGS to decouple diffuse and specular components, providing physically grounded supervision for reflective appearance and shape recovery. 
To realize this design, we adopt an enhanced 3DGS representation (\gs) as the geometric backbone and incorporate a Cubemap Encoder inspired by \dr\ to model specular reflections. 
This pBRDF-guided decomposition retains the efficiency of explicit Gaussian rendering while improving reflective shape reconstruction.

Beyond photometric cues, we also aim to exploit \emph{geometric} cues from polarization to further improve reconstruction quality. 
A natural choice is the multi-view tangent-space consistency (TSC) loss~\citep{cao2023multi,han2024nersp}. However, the existing TSC constraint is developed for implicit surface representations and relies on explicit ray-surface intersections to determine cross-view visibility masks, which is highly time-consuming. 
Due to splatting-based rasterization, this formulation cannot be directly applied to 3DGS.

To enable TSC in the Gaussian Splatting pipeline, we design a \emph{depth-guided visibility mask acquisition} mechanism that avoids ray-tracing during optimization. 
Specifically, we estimate visibility by comparing rendered depth maps with the point-to-camera distances of candidate surface points. 
This design enables, for the first time, the integration of multi-view tangent-space constraints into a 3DGS-based reconstruction framework, bridging the gap between implicit SDF-based methods and explicit Gaussian representations.

With these polarimetric constraints, \polgs resolves shape ambiguities that cannot be handled by RGB-only supervision and accurately reconstructs \textbf{\emph{reflective}} and \textbf{\emph{texture-less}} objects within $10$ minutes. 
As shown in \fref{fig.teaser} and \Tref{table:comparison}, \polgs achieves reconstruction quality comparable to state-of-the-art SDF-based methods while maintaining over $80\times$ speedup, making it suitable for real-time and large-scale reflective surface reconstruction.
\definecolor{newgreen}{rgb}{0.1,0.7,0.1}
\definecolor{3rd}{rgb}{0.95,0.95,0.65}
\definecolor{2nd}{rgb}{1,0.84,0.7}
\definecolor{1st}{rgb}{1,0.7,0.7}
\begin{table}
	\caption{Comparison of different methods in reflective surface reconstruction on real-world datasets~\citep{dave2022pandora,chen2024pisr,han2024nersp} with the image size $512*612$.}
	\label{table:comparison}
	\small
	\centering
	\resizebox{.5\textwidth}{!}{
		\begin{tabular}{lccccc}
			\toprule
			Method & Input & Reflective &Type& Accuracy & Time\\
			\midrule
			\nero & RGB Images & \color{newgreen}\ding{51} & SDF &\colorbox{1st}{high}   &  \colorbox{3rd}{8h} \\
			\mvas & Azimuths &  \color{newgreen}\ding{51} & SDF&\colorbox{2nd}{medium}   &  \colorbox{3rd}{11h} \\
			\pandora & Pol. Images&  \color{newgreen}\ding{51} & SDF&\colorbox{2nd}{medium}  & \colorbox{3rd}{10h}\\
			\nersp & Pol. Images&  \color{newgreen}\ding{51} & SDF&\colorbox{1st}{high}  & \colorbox{3rd}{10h}\\
			\pisr & Pol. Images&  \color{newgreen}\ding{51} & SDF&\colorbox{2nd}{medium}  & \colorbox{2nd}{0.5h}\\
			\gsror & RGB Images&  \color{newgreen}\ding{51} & SDF+GS&\colorbox{2nd}{medium}  & \colorbox{2nd}{1.5h}\\
			\gs & RGB Images& \color{red}\ding{55} &GS &\colorbox{3rd}{low} &  \colorbox{1st}{5m}\\
			 \dr & RGB Images&   \color{newgreen}\ding{51} &GS&  \colorbox{3rd}{low} & \colorbox{2nd}{12m} \\
			 \relight & RGB Images&  \color{newgreen}\ding{51} & GS&\colorbox{3rd}{low} & \colorbox{2nd}{25m} \\
			 \polgs~(Ours) & Pol. Images&  \color{newgreen}\ding{51} & GS&\colorbox{2nd}{medium} & \colorbox{1st}{10m} \\
			\bottomrule
			\vspace{-2.8em}	
		\end{tabular}	

	}
\end{table}


In summary, we advance the reflective surface reconstruction by proposing:
\begin{itemize}
\item \polgs, a physically-guided polarimetric 3D Gaussian Splatting framework for fast reflective surface reconstruction.

\item A pBRDF module in 3DGS that constrains both diffuse and specular components of reflective surfaces.

\item A depth-guided visibility mask acquisition mechanism that enables multi-view tangent-space consistency loss in the 3DGS framework without ray--surface intersection.
\end{itemize}

The preliminary version of this work, PolGS~\citep{han2025polgs}, considered only pBRDF integration with 3DGS and did not fully exploit polarimetric cues. 
This journal article extends PolGS~\citep{han2025polgs} by introducing a multi-view TSC constraint in 3DGS architecture and enabling it through the proposed depth-guided visibility mask acquisition strategy. 
This extension resolves azimuth ambiguity in monocular polarimetric cues and enforces stronger cross-view geometric consistency. 
Specifically, in \sref{sec:Depth-guided Visibility}, we provide a detailed analysis of the depth-guided visibility mask acquisition mechanism. 
In \sref{sec:exp}, we present more comprehensive evaluations, including expanded ablations on polarimetric information and the threshold used in the depth-guided visibility mask. We also provide broader validation on synthetic and real-world datasets, demonstrating improved robustness and reconstruction quality.

 
\section{Related work}
\label{sec:related}

Our method aims to reconstruct reflective surfaces via 3DGS using polarized images, so we summarize recent progresses in 3D reconstruction techniques, focusing on reflective surfaces through SDF-based methods, 3DGS methods, and polarized image-based reconstruction, respectively.

\paragraph{\textbf{3D reconstruction based on neural SDF representation}\quad}
Novel view synthesis has achieved great success using Neural Radiance Fields (NeRF~\citep{mildenhall2020nerf}). Motivated by the structure of the multi-layer perceptron (MLP) network within NeRFs, numerous 3D reconstruction methods have emerged that leverage implicit neural representations to predict object surfaces. Some approaches~\citep{niemeyer2020differentiable, yariv2020multiview} proposed the signed distance field (SDF) into the neural radiance field, effectively representing the surface as an implicit function. 
Other works~\citep{wang2021neus, yariv2021volume, oechsle2021unisurf, wang2022hf, wang2023neus2,li2023neuralangelo} extend it by proposing efficient framework in detailed surface reconstruction.

These methods based on SDF represent complex scenes implicitly, but they typically suffer from high computational demands and are not suitable for real-time applications. Although some methods~\citep{wang2023neus2, li2023neuralangelo} utilize hash grids and instant-NGP~\citep{mueller2022instant} structure, it is still a challenge for them to reconstruct the mesh efficiently facing the reflective surface. 

Ref-NeRF~\citep{verbin2022ref} uses the Integrated Directional Encoding (IDE) structure to estimate the specular reflection components of the object surface by using predicted roughness, view direction, and surface normals.
NeRO~\citep{liu2023nero} improves it by generating the physically-based rendering (PBR) parameters, and NeP~\citep{wang2024inverse} can better deal with the glossy surface.
TensoSDF~\citep{li2024tensosdf} combines a novel tensorial representation~\citep{chen2022tensorf} with the radiance and reflectance field for robust geometry reconstruction.
Overall, these approaches cannot avoid long optimization times, which limit their practicality for reflective scenes.

\paragraph{\textbf{3D reconstruction based on 3DGS}\quad}
3DGS~\citep{kerbl20233d} aims to address limitations of  neural radiance fields by representing complex spatial points using 3D Gaussian ellipsoids. 
However, 3D Gaussian ellipsoids often fail to conform closely to true object surfaces, leading to inaccurate geometry and degraded point-cloud quality.

To overcome these drawbacks, several extensions and modifications have been proposed. SuGaR~\citep{guedon2024sugar} approximates 2D Gaussians with
3D Gaussians, NeuSG~\citep{chen2023neusg}, GSDF~\citep{yu2024gsdf} and 3DGSR~\citep{lyu20243dgsr} integrate an extra SDF network for representing surface normals to supervise the Gaussian Splatting geometry. 
2D Gaussian Splatting~\citep{huang20242d} and Gaussian Surfels~\citep{dai2024high} have taken a different approach by transforming the 3D ellipsoids into 2D ellipses for modeling. This transformation allows for more refined constraints on depth and normal consistency, addressing the surface approximation issues more effectively. 
GOF~\citep{yu2024gaussian} achieves more realistic mesh generation through its innovative opacity rendering strategy.
PGSR~\citep{chen2024pgsr} utilizes single-view and multi-view regularization on planar Gaussians to achieve high-precision global geometric consistency.
However, these methods do not focus on reflective surface reconstruction.

\relight associates extra properties, including normal vectors, BRDF parameters,
and incident lighting from various directions to make photo-realistic relighting. 
\dr presents a deferred shading method to effectively render specular reflection with Gaussian splatting. 
\gsror integrates the SDF-based representation into 3DGS to improve surface modeling, but it still relies on a costly SDF optimization stage.
Despite these advancements, these methods still face challenges in many scenarios and cannot provide accurate geometric representations.
\vspace{-1.5em}

\paragraph{\textbf{3D reconstruction using polarized images}\quad}
Polarized images are widely used in Shape from Polarization (SfP)~\citep{miyazaki2003polarization, baek2018simultaneous, smith2018height, deschaintre2021deep, lei2022shape, ba2020deep, lyu2023shape,li2024deep, yang2024gnerp, lyu2024sfpuel}, reflection removal~\citep{li2020reflection,lyu2022physics,wang2025polarized}, and some downstream tasks~\citep{zhou2023polarization, li2023polarized} due to the strong physics-preliminary information in the Stokes field. The SfP task aims to predict the surface normal captured by the polarization camera under the single distant light~\citep{lyu2023shape, smith2018height} or unknown ambient light~\citep{ba2020deep,lei2022shape}. Multi-view 3D reconstruction works using polarized images~\citep{zhao2022polarimetric} try to settle down the $\pi$ and $\pi/2$ ambiguities with the Angle of Polarization (AoP). \pandora first uses polarized images in neural 3D reconstruction work, following the relevant constraints of pBRDF~\citep{baek2018simultaneous}. \mvas leverages multi-view AoP maps to generate tangent spaces for surface points during the optimization process, which can 
reconstruct mesh without rendering supervision. \nersp combines the photometric and geometric cues from polarized images and generates better results under sparse views for reflective surfaces. 
PISR~\citep{chen2024pisr} focuses on texture-less specular surface 
and integrates the multi-resolution hash grid for efficiency.
NeISF~\citep{li2024neisf} considers inter-reflection and models multi-bounce polarized light paths during rendering. Furthermore, NeISF++~\citep{li2025neisf++} extends this approach to conductive surfaces. 
Despite these advancements, computational cost remains a significant limitation for many polarized-based 3D reconstruction methods.
 \section{Preliminaries}
\label{sec:prelim}
\subsection{Gaussian Surfels model}
Our method adopts \gs as the base framework, leveraging its superior geometric representation capabilities. 
Following \gs, we use a set of unstructured Gaussian kernels $\{G_i=\{\textbf{x}_i,\textbf{s}_i,\textbf{r}_i,o_i,C_i\}|i\in
 \mathcal{N}\}$ to represent the structure of 3DGS, where  $\textbf{x}_i\in\mathbb{R}^3$ denotes the center 
 position of each Gaussian kernel, $\textbf{s}_i=[s^x_i,s^y_i,0]^\top\in\mathbb{R}^3$ is the scaling factors of $x$ and 
 $y$ axes after flatting the 3D Gaussians~\citep{kerbl20233d}, $\textbf{r}_i\in\mathbb{R}^4$ is the rotation quaternion, 
 $o_i\in\mathbb{R}$ is the opacity, and $C_i\in\mathbb{R}^k$ represents the spherical harmonic coefficients of each Gaussian. 
 The Gaussian distribution can be defined by the covariance matrix $\Sigma$ of a 3D Gaussian as:
\begin{equation}
    G(\textbf{x};\textbf{x}_i,\Sigma_i)=\text{exp}(-\frac{1}{2}(\textbf{x}-\textbf{x}_i)^\top\Sigma_i^{-1}(\textbf{x}-\textbf{x}_i)),
\end{equation}
where $\Sigma_i$ can be represented as:
\begin{equation}
\begin{aligned}
	\label{equation:sigma}
    \Sigma_i &= \textbf{R}(\textbf{r}_i)\textbf{s}_i\textbf{s}_i^\top\textbf{R}(\textbf{r}_i)^\top\\
    &= \textbf{R}(\textbf{r}_i)\text{Diag}[(s^x_i)^2,(s^y_i)^2,0]\textbf{R}(\textbf{r}_i)^\top,
\end{aligned}
\end{equation}
where Diag[·] indicates a diagonal matrix and $\textbf{R}(\textbf{r}_i)$ is a $3 \times 3$
rotation matrix represented by $\textbf{r}_i$.

\paragraph{Gaussian splatting\quad}
The novel view rendering process can be represented as~\citep{kerbl20233d}: 
\begin{equation}
    C=\sum_{i=0}^nT_i\alpha_ic_i,
\label{eq:gs_render}
\end{equation}
where $T_i=\prod_{j=0}^{i-1}(1-\alpha_j)$ is the transmittance, $\alpha_i = G'(\textbf{u};\textbf{u}_i,\Sigma'_i)o_i$ 
is alpha-blending weight, which is the product of opacity and the Gaussian weight based on pixel $\textbf{u}$. 
In order to speed up the rendering process, the 3D Gaussian in \eref{equation:sigma} 
is re-parameterized in 2D ray space~\citep{zwicker2002ewa} as $G'$:
\begin{equation}
    G'(\textbf{u};\textbf{u}_i,\Sigma'_i)=\text{exp}(-\frac{1}{2}(\textbf{u}-\textbf{u}_i)^\top {\Sigma'_i}^{-1}(\textbf{u}-\textbf{u}_i)),
\end{equation}
where ${\Sigma'_i}= (\textbf{J}_k\textbf{W}_k\Sigma_i\textbf{W}_k^\top\textbf{J}_k^\top)[:2,:2]$ represents the covariance matrix in the 2D ray space. The $\textbf{W}_k$
is a viewing transformation matrix for input image $k$ and $\textbf{J}_k$ is the affine approximation of the projective transformation.

The depth $\tilde{\textbf{D}}$ and normal $\tilde{\textbf{N}}$ for each pixel can also be
calculated via Gaussian splatting and alpha-blending~\citep{dai2024high}:
\begin{equation}
        \tilde{\textbf{D}}=\frac{1}{1-T_{n+1}}\sum_{i=0}^{n}T_i\alpha_id_i(\textbf{u}),
\end{equation}
\begin{equation}
        \tilde{\textbf{N}}=\frac{1}{1-T_{n+1}}\sum_{i=0}^{n}T_i\alpha_i\textbf{R}_i[:,2].
\label{eq:gs_normal}
\end{equation}
Specifically, according to~\gs, the depth of pixel $\textbf{u}$ for each Gaussian kernel $i$
is computed by calculating the intersection of the ray cast through pixel $\textbf{u}$ 
with the Gaussian ellipse during splatting. So $d_i(\textbf{u})$ can be represented by Taylor expansion as:
\begin{equation}
    d_i(\textbf{u}) = d_i(\textbf{u}_i) + (\textbf{W}_k\textbf{R}_i)[2,:]\textbf{J}_{pr}^{-1}(\textbf{u}-\textbf{u}_i),
\end{equation}
where $\textbf{J}_{pr}^{-1}$ is the Jacobian inverse mapping one pixel from image space to tangent plane of the Gaussian surfel~\citep{zwicker2001surface}, and $(\textbf{W}_k\textbf{R}_i)$ transforms the rotation matrix of a Gaussian surfel
to the camera space.

\subsection{Polarimetric image formation model}
\label{sec:polar image formation}
Polarization cameras can capture polarized images in a single shot, 
which can be represented as four different polarized angles of images 
$I=[I_0,I_{45},I_{90},I_{135}]$. 
The Stokes vector $\V{s} = \{s_0,s_1,s_2,s_3\}$ can be calculated by:
\begin{equation}
    \V{s} = \begin{bmatrix}\frac{1}{2}
        (I_0+I_{45}+I_{90}+I_{135})\\
        I_{0}-I_{90}\\
        I_{45}-I_{135}\\
        0
    \end{bmatrix},
\end{equation}
where we assume the light source is not circularly polarized following prior works~
\citep{dave2022pandora,han2024nersp},
thus the $s_3$ is 0. The Stokes vector can be used to compute the angle of polarization~(AoP), \ie
\begin{eqnarray}
	\varphi = \frac{1}{2}\arctan(\frac{s_2}{s_1}).
\end{eqnarray}

According to \pandora and \nersp, we assume the incident environmental illumination is unpolarized, and the Stokes vector for the incident light direction $\bm{\omega}$ can be expressed as:
\begin{equation}
    \V{s}_i(\bm{\omega}) = L(\bm{\omega}) [1, 0, 0, 0]^\top,
    \label{eq:env_stok}
\end{equation}
where $L(\bm{\omega})$ represents the light intensity. The polarization camera captures outgoing light that becomes partially polarized due to 
reflection, which is modeled using a $4 \times 4$ Muller matrix $\V{H}$. 
The outgoing Stokes vector $\V{s}$ then is formulated as the integral of the incident Stokes vector multiplied by this Mueller matrix:
\begin{equation}
    \V{s}(\V{v}) = \int_{\Omega} \V{H} \V{s}_i(\bm{\omega}) \,\, d\bm{\omega},
    \label{eq:out_stok}
\end{equation}
where $\V{v}$ indicates the view direction and $\Omega$ is the integral domain. 
Following the polarized BRDF (pBRDF) model proposed by~\cite{baek2018simultaneous}, the output Stokes vector can be divided into diffuse and specular components, 
represented by $\V{H}_d$ and $\V{H}_s$ respectively. $\V{s}$ 
can be represented as:
\begin{equation}
    \V{s}(\V{v}) = \int_{\Omega} \V{H}_d \V{s}_i(\bm{\omega}) \,\, d\bm{\omega} + \int_{\Omega} \V{H}_s \V{s}_i(\bm{\omega}) \,\, d\bm{\omega}.
    \label{eq:decomp_stok}
\end{equation}

Based on the derivations from \pandora and \nersp, The output Stokes vector can be 
further specified as:
\begin{eqnarray}
    \V{s}(\V{v}) =   L_d \begin{bmatrix}
        T_o^+ \\
        T_o^- \cos(2 \phi_n) \\
        - T_o^- \sin(2 \phi_n) \\
        0
    \end{bmatrix} +  L_s \begin{bmatrix}
        R^+ \\
        R^- \cos(2 \phi_h) \\
        -R^- \sin(2 \phi_h) \\
        0
    \end{bmatrix},
    \label{eq:photometric_cue}
\end{eqnarray}
where $L_d = \int_{\Omega} \rho L(\bm{\omega})  \bm{\omega}^\top \V{n} \, T_i^+T_i^-   \, d\bm{\omega}$ 
denotes the diffuse radiance associated with the surface normal $\V{n}$, 
Fresnel transmission coefficients~\citep{baek2018simultaneous} $T_{i, o}^+$ and 
$T_{i, o}^-$. The diffuse albedo is represented by $\rho$, and $\phi_n$
is the azimuth angle of the incident light. 

Similarly, $L_s = \int_{\Omega} L(\bm{\omega})  \frac{DG}{4 \V{n}^\top\V{v}}   \,\, d\bm{\omega}$ 
denotes the specular radiance, which involves Fresnel reflection coefficients~\citep{baek2018simultaneous} 
$R^+$ and $R^-$, and the incident azimuth angle $\phi_h$ concerning the half vector 
$\V{h} = \frac{\bm{\omega} + \V{v}}{\|\bm{\omega} + \V{v}\|_2^2}$. For the sake of simplification in our model, we assume that $\phi_h$
is equivalent to $\phi_n$. The Microfacet model incorporates the normal distribution and shadowing terms represented by 
$D$ and $G$~\citep{walter2007microfacet}. 

\subsection{Multi-view tangent space consistency with AoP}
Given AoP $\varphi$ (as mentioned in \sref{sec:polar image formation}), the azimuth angle of the surface can be either $\varphi + \pi /2$ or $\varphi + \pi$, known as the $\pi$ and $\pi / 2$ ambiguity, where $\pi / 2$ ambiguity depending on whether the surface is specular or diffuse dominant.

Following \mvas, for a scene point $\V{x}$, its surface normal $\V{n}$ and the projected azimuth angle $\phi$ in one camera view follow the relationship as
\begin{eqnarray}
	\V{r}_1^\top \V{n} \cos\phi - \V{r}_2^\top \V{n} \sin\phi = 0,
	\label{eq:azi_rotation}
\end{eqnarray}
where $\V{R} = [\V{r}_1, \V{r}_2, \V{r}_3]^\top$ is the rotation matrix of the camera pose. We can further  re-arrange \eref{eq:azi_rotation} to get the orthogonal relationship between surface normal and a projected tangent vector $\V{t}(\phi)$ as defined below,
\begin{eqnarray}
	\V{n}^\top \underbrace{(\cos\phi \V{r}_1 -  \sin\phi \V{r}_2 )}_{\V{t}(\phi)} = 0.
	\label{eq:projected_tan}
\end{eqnarray}
The $\pi$ ambiguity between AoP and azimuth angle can be naturally resolved as \eref{eq:projected_tan} stands if we add $\phi$ by $\pi$. The $\pi/2$ ambiguity can be addressed by using a pseudo projected tangent vector $\hat{\V{t}}(\phi)$ such that
\begin{eqnarray}
	\V{n}^\top \underbrace{(\sin\phi \V{r}_1  +  \cos\phi \V{r}_2 )}_{\hat{\V{t}}(\phi)} = 0.
	\label{eq:projected_tan_psu}
\end{eqnarray}
If one scene point $\V{x}$ is observed by $k$ views, we can stack \eref{eq:projected_tan} and \eref{eq:projected_tan_psu} based on $k$ different rotations and observed AoPs, leading to a linear system 
\begin{eqnarray}
	\V{T}(\V{x}) \V{n}(\V{x}) = \V{0}.
	\label{eq:tangent_consistency}
\end{eqnarray}

In \mvas and \nersp, the above linear system is used to formulate a multi-view tangent space consistency loss to constrain the surface normal estimation in neural implicit representations. However, directly applying such constraints in Gaussian Splatting frameworks remains challenging due to the lack of an explicit visibility definition across views. In \sref{sec:Depth-guided Visibility}, we propose a depth-guided visibility mask acquisition mechanism to address it.
\begin{figure}[t]
	\centering
	\includegraphics[width=\linewidth]{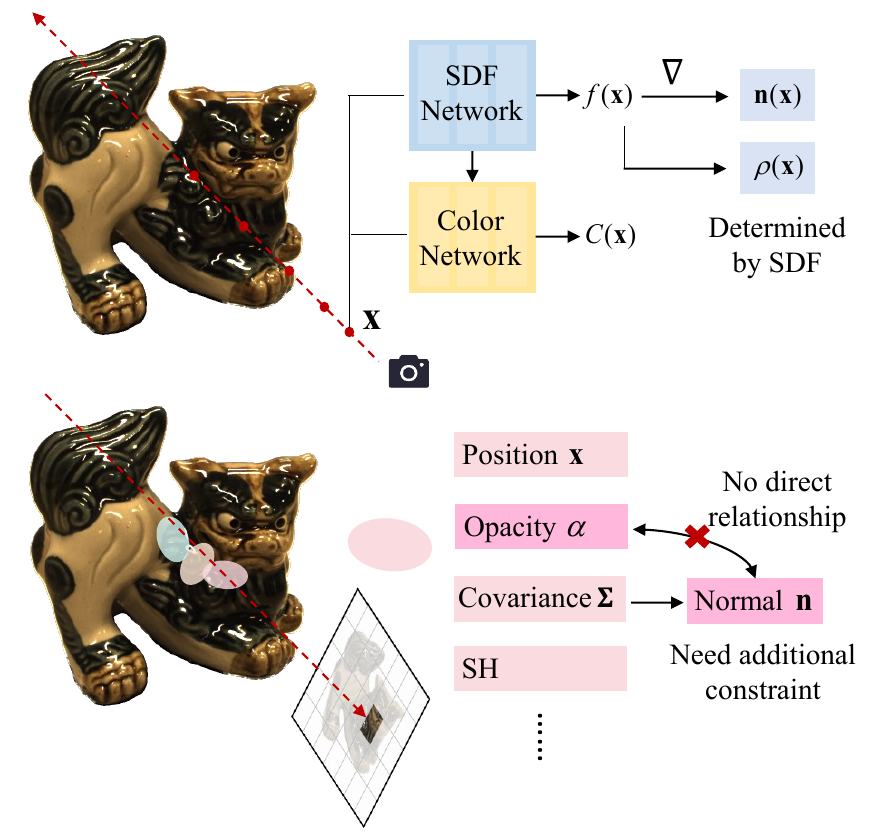}
	\caption{Comparison between SDF-based and 3DGS-based method on geometry representation.  (\textbf{Top}) The surface normal of a point has a strong relationship with its opacity in NeuS~\citep{wang2021neus}. (\textbf{Bottom}) The surface normal of a point has no direct dependence on its opacity in \gs.}
	\label{fig:normal-analysis}
    \vspace{-1em}
\end{figure}
\vspace{-2em}
\section{Proposed method}

\label{sec:method}
The \polgs framework integrates multi-view polarized images, their corresponding masks, and camera pose information to produce a rich output that includes diffuse and specular components represented as Stokes vectors across various views, a reconstructed geometric mesh, and estimated environment light. In this section, we will analyze the surface normal representations employed in SDF-based approaches and 3DGS methods. Then we introduce a theoretical foundation for our polarization-guided reconstruction pipeline by using the polarimetric image formation model and multi-view tangent space consistency constraint.
\vspace{-1em}
\begin{figure*}
	\centering
	\includegraphics[width=\linewidth]{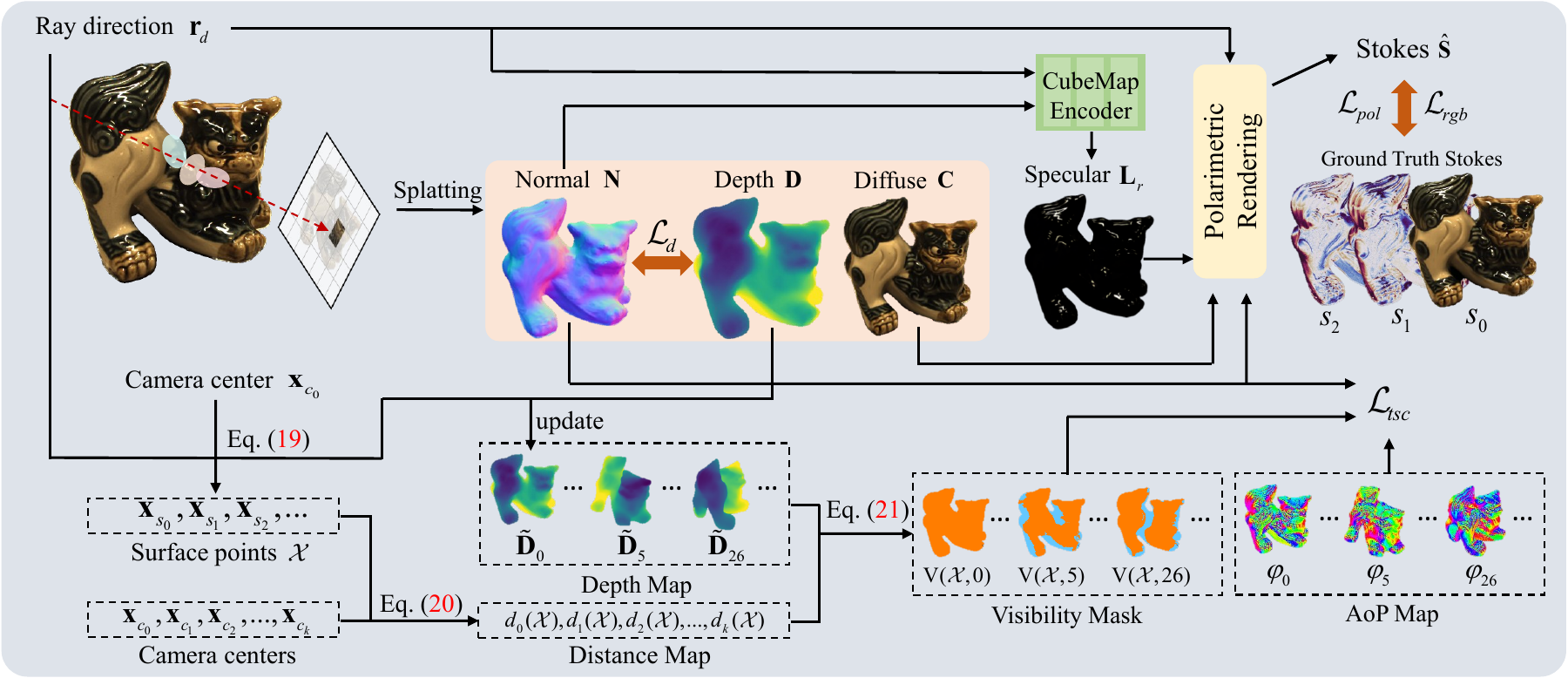}
	\caption{Pipeline of \polgs. We re-rendered Stokes vectors $\hat{s}$ by using the diffuse color $C$ from 3DGS and specular color $L_r$ from Cubemap encoder module, which is supervised by the ground truth Stokes information. Tangent-space consistency constraint is applied to the surface normal estimated by 3DGS with the depth-guided visibility mask and AoP information.} 
	\label{fig:pipeline}
    \vspace{-1em}
\end{figure*}
\subsection{Analysis of surface normal representation}
In implicit neural networks, researchers employ signed distance functions (SDFs) to define surface boundaries, where  
 surface normals are derived from the gradients of SDF. However, 3DGS methods often treat the surface normal as inherent properties typically.
In this section, we provide a comparative analysis of surface normal representations in SDF-based methods versus 3DGS, explaining the reason we use polarimetric cues in the Gaussian Splatting method.

\vspace{-0.5em}
\paragraph{SDF-based representation\quad}
The implicit neural network approach, exemplified by NeuS~\citep{wang2021neus}, enhances the constraints on the reconstructed surface by incorporating a Signed Distance Function (SDF). As illustrated at the top of \fref{fig:normal-analysis}, the SDF value \( f(\textbf{x}) \), generated by the SDF Network, can be used to compute the normal vector \( \textbf{n}(\textbf{x}) \) at point \( \textbf{x} \). According to the definition in \citep{wang2021neus}, the opacity \( \rho(\textbf{x}) \)   is given by:
\begin{equation}
    \rho(\textbf{x}) = \max\left(\frac{-\frac{d\Phi_s}{dx}(f(\textbf{x}))}{\Phi_s(f(\textbf{x}))}, 0\right),
\label{eq:neus_density}
\end{equation}
where \( \Phi_s \) is a sigmoid function.

During optimization, the surface normal and the opacity at a given point influence each other. Consequently, after optimization, points farther from the surface tend to have lower opacity values, while the final normal vector is predominantly determined by points located on the surface. This mutual interaction results in a more precise and accurate representation of the surface.
\vspace{-0.5em}
\paragraph{3DGS-based representation\quad}
In the case of \gs, surface normal is derived from the vector perpendicular to the plane of a 2D ellipsoid. As shown at the bottom of \fref{fig:normal-analysis}, for a single Gaussian kernel, there is no inherent relationship between opacity and the surface normal. 
The pixel-level constraint in 3DGS creates a probabilistic representation where no single Gaussian kernel is definitively assigned to a surface. Consequently, multiple Gaussian kernel configurations can potentially represent equivalent surface geometries, introducing inherent reconstruction ambiguity. 
The optimization mechanism effectively blends contributions from multiple Gaussian points, which offers reconstruction adaptability but compromises the precision of individual point geometric characterizations.

It is obvious that using the normal prediction model~\citep{eftekhar2021omnidata} as a prior constraint in Gaussian Splatting benefits object reconstruction, as demonstrated by \gs. However, the normal prediction model is not always reliable especially in reflective cases, highlighting the need for a more robust method to provide prior information.
\vspace{-0.8em}
\paragraph{Polarimetric information for reflective surfaces\quad}
Reflective surface reconstruction is challenging due to the view-dependent appearance. Unlike traditional RGB image inputs, polarimetric information can effectively constrain the surface normal during the rendering of Stokes vectors with the pBRDF model~\citep{baek2018simultaneous} and the multi-view tangent space consistency constraint.
Leveraging this property, we incorporate polarimetric information as a prior in shape reconstruction and use the 3DGS-based method to accelerate this process.

\subsection{Pipeline of \polgs}
\label{sec:pipeline}

\paragraph{Network structure\quad}
The network structure of \polgs is shown in \fref{fig:pipeline}.
The framework comprises two primary components: the Gaussian Surfels module and the CubeMap Encoder module. 
Initially, the Gaussian Surfels module is employed to estimate the diffuse component of the object, similar to the original Gaussian Surfels framework. Subsequently, we utilize the CubeMap Encoder 
to assess the specular component, akin to the approach taken in \dr. While the CubeMap Encoder does not provide the roughness component, it effectively handles reflective or rough surfaces and maintains high computational efficiency due to its CUDA-based implementation. 

To enhance the rendering process, we incorporate a pBRDF model into the rendering formulation. 
This addition introduces a polarimetric constraint that further refines the Gaussian Splatting method, 
enabling more accurate and realistic 3D reconstructions.
The final rendering formulation model following 
\eref{eq:photometric_cue}
can be represented as:
\begin{equation}
    \hat{\V{s}} = \begin{bmatrix}
        \hat{s}_0 \\
        \hat{s}_1 \\
        \hat{s}_2 \\
        0
    \end{bmatrix}=C \begin{bmatrix}
        T_o^+ \\
        T_o^- \cos(2 \phi_n) \\
        - T_o^- \sin(2 \phi_n) \\
        0
    \end{bmatrix}+ L_r\begin{bmatrix}
        R^+ \\
        R^- \cos(2 \phi_h) \\
        -R^- \sin(2 \phi_h) \\
        0
    \end{bmatrix},
\label{eq:final_stokes}
\end{equation}
where $C$ is the diffuse color after Gaussian Surfels rendering according to \eref{eq:gs_render} and 
$L_r$ is the specular color after CubeMap rendering $E(\cdot)$, which can be represented as 
$L_r = E(\textbf{r}_d-2(\textbf{r}_d\cdot \textbf{n})\textbf{n})$.

\begin{figure}[h]
    \centering
    \includegraphics[width=\linewidth]{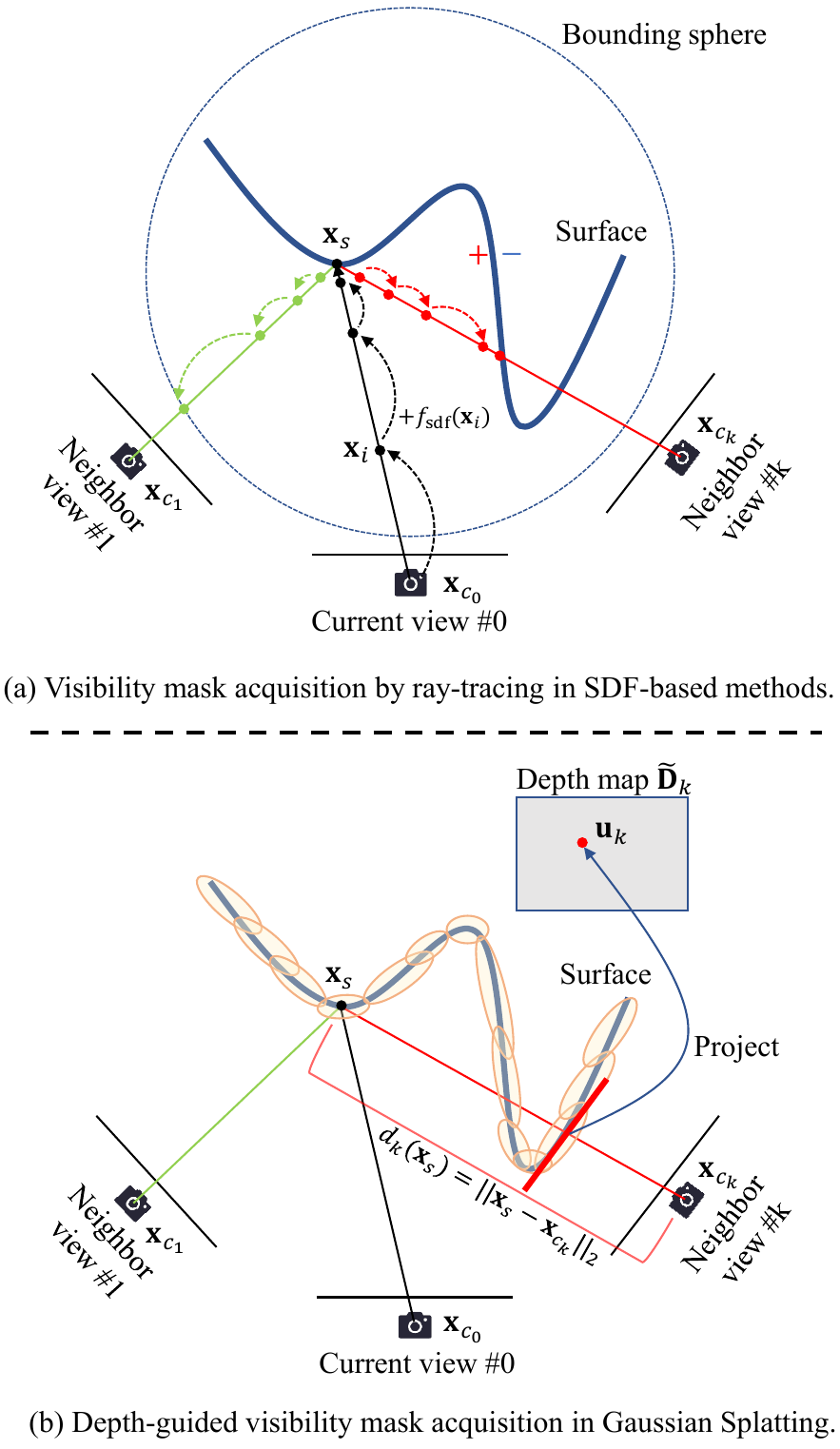}
    \caption{Comparison of visibility mask acquisition strategies between SDF-based methods and our method based on Gaussian Splatting.}
    \label{fig:visibility}
    \vspace{-0.5em}
\end{figure}

\paragraph{Depth-guided visibility mask\quad}
\label{sec:Depth-guided Visibility}
In addition to the polarimetric rendering, \polgs employs multi-view tangent space consistency to constrain surface geometry~\citep{cao2023multi,han2024nersp}. However, a key prerequisite is a reliable visibility mask to indicate whether a surface point is observable from a neighboring view.
In implicit SDF-based representations, visibility can be classically evaluated via ray-tracing with the signed distance field (SDF). As shown in \fref{fig:visibility}~(a), a surface point \( \textbf{x}_s \) can be confirmed along the ray originating at camera center $\textbf{x}_{c_0}$ with direction $r_d$ by iteratively stepping the ray by the SDF value until the first zero-crossing. To determine the visibility of $\textbf{x}_s$ from another view with camera center $\textbf{x}_{c_k}$, one traces the ray from $\textbf{x}_s$ to $\textbf{x}_{c_k}$ and checks whether any surface intersection occurs before exiting the bounding sphere. If no intersection is detected, the visibility mask from view $k$ is set as $\textbf{V}(\textbf{x}_s, k) = 1$; otherwise, $\textbf{V}(\textbf{x}_s, k) = 0$.

However, this strategy is not directly applicable to Gaussian Splatting frameworks because they lack an explicit ray--surface intersection test. In 3DGS-based methods, the surface is represented by a set of Gaussian surfels and rendered via splatting and alpha compositing, making it non-trivial to backtrace whether a surface point is visible from another view.
To address this limitation, we propose a \emph{depth-guided visibility mask} acquisition strategy that leverages the depth maps produced during Gaussian Splatting.

As shown in \fref{fig:visibility}~(b), we construct a set of pseudo-surface points
\(\mathcal{X}=\{\mathbf{x}_{s}\in\mathbb{R}^3\}\) by back-projecting the rendered depth map of the reference view \(0\):
\begin{equation}
\mathbf{x}_{s}(\mathbf{u}) = \mathbf{x}_{c_0} + \tilde{\textbf{D}}_0(\mathbf{u})\,\mathbf{r}_0(\mathbf{u}),
\end{equation}
where \(\mathbf{x}_{c_0}\) is the camera center of the reference view, \(\tilde{\textbf{D}}_0\) denotes the rendered depth map,
\(\mathbf{u}\) is the pixel coordinate, and \(\mathbf{r}_0(\mathbf{u})\) is the corresponding unit ray direction in world coordinates.

For a neighboring view \(k\) with camera center \(\mathbf{x}_{c_k}\), we define the geometric distance from the camera to \(\mathbf{x}_{s}\) as
\begin{equation}
d_k(\mathbf{x}_{s}) = \| \mathbf{x}_{s} - \mathbf{x}_{c_k} \|_2.
\end{equation}

Meanwhile, \polgs renders a depth map \(\tilde{\textbf{D}}_k\) for view \(k\) at each splatting iteration.
We cache \(\tilde{\textbf{D}}_k\) in an auxiliary buffer and detach it from backpropagation.
By projecting \(\mathbf{x}_{s}\) onto the image plane of view \(k\), we obtain the corresponding pixel coordinate
\(\mathbf{u}_k = \pi_k(\mathbf{x}_{s})\) and query the rendered depth value \(\tilde{\textbf{D}}_k(\mathbf{u}_k)\).

Ideally, if \(\mathbf{x}_{s}\) is visible from view \(k\), the rendered depth along the viewing ray should agree with the geometric distance \(d_k(\mathbf{x}_{s})\).
Therefore, we define the visibility mask as
\begin{equation}\label{eq:visibility}
\mathbf{V}(\mathbf{x}_{s}, k) =
\begin{cases}
1, & \left| \tilde{\textbf{D}}_k\!\big(\pi_k(\mathbf{x}_{s})\big) - d_k(\mathbf{x}_{s}) \right| < \tau, \\
0, & \text{otherwise},
\end{cases}
\end{equation}
where \(\tau\) is a tolerance threshold that accounts for depth uncertainty caused by Gaussian blending and discretization effects.

\begin{figure}[t]
	\centering
	\includegraphics[width=\linewidth]{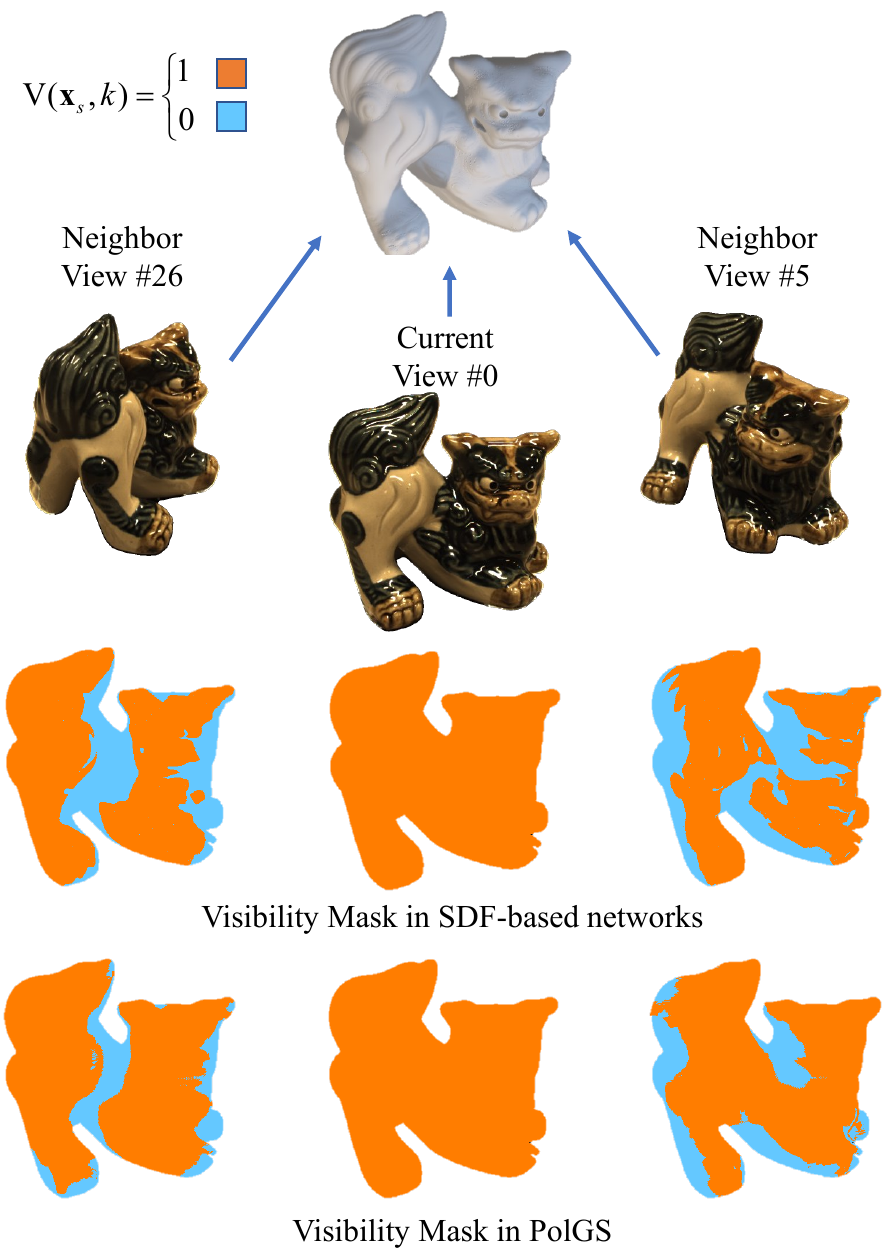}
	\caption{Visualization of visibility masks of one current view in SDF-based networks and in \polgs.}
	\label{fig:visi_mask}
\end{figure}

\Fref{fig:visi_mask} compares visibility masks of SDF-based networks and \polgs from one current view. In SDF-based methods, we use ground-truth normals to supervise and obtain the visibility mask. Our depth-guided visibility mask estimation yields comparable results without explicit ray-tracing. Building upon this, we can formulate our multi-view tangent-space consistency constraint directly corresponding to the defined linear system. Details of the ablation study about the visibility mask will be discussed in \sref{exp:visibility}.


\subsection{Training}
The overall training loss in \polgs formulated as a comprehensive weighted sum of multiple loss components:
\begin{equation}
	\mathcal{L} = \mathcal{L}_{rgb}+\lambda_{1}\mathcal{L}_{pol}+ \lambda_{2}\mathcal{L}_{tsc}+\lambda_{3}\mathcal{L}_{m}+\lambda_{4}\mathcal{L}_{o}	+\lambda_{5}\mathcal{L}_{d},
\end{equation}
where we set $\lambda_{1}=1$, $\lambda_{2}=0.1$, $\lambda_{3}=0.1$, $\lambda_{4}=0.01$, $\lambda_{5}=0.01+0.1\cdot (iteration/15000)$ to balance the loss function.
\vspace{-1em}

\paragraph{Rendering Stokes loss $\mathcal{L}_{rgb}$ and $\mathcal{L}_{pol}$\quad}
The rendering Stokes loss is combined with the $s_0$ (unpolarized image) rendering loss as 3DGS~\citep{kerbl20233d} and the 
$s_1$, $s_2$ rendering loss as \pandora. These two loss functions can be represented as:
\begin{align}
        &\mathcal{L}_{rgb} = 0.8\cdot L_{1} (s_0,\hat{s}_0)+0.2\cdot L_{D-SSIM}(s_0,\hat{s}_0),\\
        &\mathcal{L}_{pol} =L_{1} (s_1,\hat{s}_1)+L_{1} (s_2,\hat{s}_2).
\end{align}

\paragraph{Tangent-space consistency loss $\mathcal{L}_{\text{tsc}}$\quad}
The tangent-space consistency loss aims to strengthen the constraints of multi-view polarization information on surface normal estimation.
It ensures that the tangent plane consistency constraint is applied only when surface points are visible from the corresponding views. This is achieved through depth-guided visibility mask acquisition.
It is defined as follows:
\begin{equation}
\mathcal{L}_{\text{tsc}}
=
\sum_{\mathbf{x}_{s} \in \mathcal{X}}
\sum_{k \in K}
\mathbf{V}(\mathbf{x}_{s}, k)
\,
\left\|
\mathbf{T}_k(\mathbf{x}_{s})\,\mathbf{n}(\mathbf{x}_{s})
\right\|_2^2,
\end{equation}
where $\mathcal{X}$ denotes the set of surface points, and $K$ represents the number of neighboring views.

\paragraph{Mask loss $\mathcal{L}_{m}$}\quad
The mask loss is used to make the rendering results of the object more accurate, which can be represented as:
\begin{equation}
    \mathcal{L}_{m} = \Sigma\text{BCE}(\textbf{M},\hat{\textbf{M}}).
\end{equation}

\paragraph{Opacity loss $\mathcal{L}_{o}$}\quad
The opacity loss follows \gs to encourage the opacity of the Gaussian points to be close to 1 or 0. It can be represented as:
\begin{equation}
    \mathcal{L}_{o} = \Sigma\text{exp}(-{20(o_i-0.5)}^2).
\end{equation}

\paragraph{Depth-normal consistency loss $\mathcal{L}_{d}$}\quad
The depth-normal consistency loss follows \gs to make the rendered depth and normal of the object to be more consistent.
It can be represented as:
\begin{equation}
    \mathcal{L}_{d} = 1-\tilde{\textbf{N}}\cdot N(V(\tilde{\textbf{D}})),
\end{equation}
where $V(\cdot)$ transforms each pixel and its depth to a 3D point and $N(\cdot)$ calculates the normal from neighboring points using the cross product.
 
\setlength{\aboverulesep}{-0.7pt}
\setlength{\belowrulesep}{-0pt}
\begin{table*}
	\caption{Comparisons on synthetic dataset (\smvp) evaluated by mean angular error (MAE)~($\downarrow$) with degree and Chamfer distance (CD)~($\downarrow$) in millimeter (mm), respectively. 
	Best and second results in SDF-based and 3DGS-based methods (except PolGS~\citep{han2025polgs}) are highlighted as \colorbox[rgb]{1,0.7,0.7}{1st} and \colorbox[rgb]{1,1,0.4}{2nd}. The time consumed is shown on the far right side of the table.
	}
	\label{table:shape_quantitative_smvp}
	\vspace{-0.5em}
	\centering
	\resizebox{\textwidth}{!}{
\begin{tabular}{lcccccccc|cc|c}
	\toprule
	 & \multicolumn{2}{c}{\sc Hedgehog } & \multicolumn{2}{c}{\sc Squirrel} & \multicolumn{2}{c}{\sc Snail} & \multicolumn{2}{c|}{\sc David} & \multicolumn{2}{c|}{Mean}& \\
	\cmidrule{2-11}
		\multicolumn{1}{c}{\multirow{-2}{*}{Method}} & MAE & CD & MAE & CD & MAE & CD & MAE & CD & MAE & CD& \multicolumn{1}{c}{\multirow{-2}{*}{time}}\\
	\midrule
		\pandora&9.41&9.50  &10.85& 5.88&8.08&10.97&14.75& 4.88&10.77&7.81&10 h\\
	 
	 \nersp& 8.94&6.57 &8.23&  \colorbox[rgb]{1,1,0.4}{3.02}&5.56& \colorbox[rgb]{1,1,0.4}{3.72} &15.38&4.18 &9.53&\colorbox[rgb]{1,1,0.4}{4.37}&10 h\\
	
	 \mvas&\colorbox[rgb]{1,1,0.4}{4.30}&\colorbox[rgb]{1,1,0.4}{4.22} & \colorbox[rgb]{1,1,0.4}{6.10}& 3.73 & \colorbox[rgb]{1,1,0.4}{3.30}& 7.87 & \colorbox[rgb]{1,1,0.4}{8.47}& \colorbox[rgb]{1,1,0.4}{3.21}&\colorbox[rgb]{1,1,0.4}{5.54}&4.76&11 h\\
	 	\gsror &9.72& 4.45&9.25& 4.20&7.24& 5.16&14.02&4.14&10.06&4.49& 1.5 h \\
    
	 \nero &\colorbox[rgb]{1,0.7,0.7}{3.40}&\colorbox[rgb]{1,0.7,0.7}{3.69} & \colorbox[rgb]{1,0.7,0.7}{3.55}& \colorbox[rgb]{1,0.7,0.7}{1.86} & \colorbox[rgb]{1,0.7,0.7}{2.67}& \colorbox[rgb]{1,0.7,0.7}{3.71} &\colorbox[rgb]{1,0.7,0.7}{7.64}& \colorbox[rgb]{1,0.7,0.7}{2.88} &\colorbox[rgb]{1,0.7,0.7}{4.32}&\colorbox[rgb]{1,0.7,0.7}{3.04}&8 h\\

    \midrule
    
    \dr &\colorbox[rgb]{1,1,0.4}{12.28}& 12.66&17.18& 11.20 &11.42& 20.70&20.56&7.91&15.36&13.12& 12 min \\

    \relight &14.40& 19.84&\colorbox[rgb]{1,1,0.4}{15.41}&18.97 &\colorbox[rgb]{1,1,0.4}{9.08}& 19.04&\colorbox[rgb]{1,1,0.4}{15.47}& 13.47&\colorbox[rgb]{1,1,0.4}{13.59}&17.83& 25 min \\
	
	\gs &16.50&\colorbox[rgb]{1,1,0.4}{8.82}&21.65& \colorbox[rgb]{1,1,0.4}{9.53}&19.05& \colorbox[rgb]{1,1,0.4}{14.04}&21.56& \colorbox[rgb]{1,1,0.4}{7.39}&19.69&\colorbox[rgb]{1,1,0.4}{9.95}&5 min  \\
    
\polgs~(Ours) &\colorbox[rgb]{1,0.7,0.7}{8.33}&\colorbox[rgb]{1,0.7,0.7}{6.99}&\colorbox[rgb]{1,0.7,0.7}{9.18}& \colorbox[rgb]{1,0.7,0.7}{6.08}&\colorbox[rgb]{1,0.7,0.7}{7.25}&\colorbox[rgb]{1,0.7,0.7}{9.20}&\colorbox[rgb]{1,0.7,0.7}{13.36}& \colorbox[rgb]{1,0.7,0.7}{5.12}&\colorbox[rgb]{1,0.7,0.7}{9.53}&\colorbox[rgb]{1,0.7,0.7}{6.85}&10 min\\				
\midrule
PolGS~\citep{han2025polgs} w/o $\mathcal{L}_{pol}$  & 11.39 & 7.97 & 11.83 & 6.39 & 9.82 & 10.75 & 15.44 & 5.90 & 12.12 & 7.75 & 7 min \\
PolGS~\citep{han2025polgs} &10.83& 7.62&11.42& 6.28 &9.64& 10.85&13.99&5.30&11.47&7.51& 7 min \\

    \bottomrule
\end{tabular}	
}
\end{table*}
 
\begin{table*}
	\caption{Comparisons on real-world datasets (\rmvp and \pisr) evaluated by mean angular error (MAE)~($\downarrow$) with degree and Chamfer distance (CD)~($\downarrow$) in millimeter (mm), respectively. 
	Best and second results in SDF-based and 3DGS-based methods (except PolGS~\citep{han2025polgs}) are highlighted as \colorbox[rgb]{1,0.7,0.7}{1st} and \colorbox[rgb]{1,1,0.4}{2nd}. The time consumed is shown on the far right side of the table.
	}
	\label{table:shape_quantitative_rmvp_pisr}
	\vspace{-0.5em}
	\centering
	\normalsize
	\resizebox{\textwidth}{!}{
\begin{tabular}{lcccc|cccc|cc|c}
	\toprule
	& \multicolumn{4}{c|}{\rmvp}& \multicolumn{4}{c|}{\pisr}&& \\
	\cmidrule{2-9}
	 & \multicolumn{2}{c}{\sc Shisa} & \multicolumn{2}{c|}{\sc Frog} & \multicolumn{2}{c}{\sc L-Rabbit} & \multicolumn{2}{c|}{\sc S-Rabbit}&\multicolumn{2}{c|}{\multirow{-2}{*}{Mean}}& \\
	\cmidrule{2-11}
		\multicolumn{1}{c}{\multirow{-3}{*}{Method}} & MAE & CD & MAE & CD & MAE & CD & MAE & CD& MAE & CD& \multicolumn{1}{c}{\multirow{-3}{*}{time}}\\
	\midrule
		\pandora&12.93&11.29&15.86&7.88&13.47&16.53&\colorbox[rgb]{1,1,0.4}{11.45}&15.39&13.43&12.77&10 h\\
	 
	 \nersp& 10.79& 7.39& 15.62& 6.68& 13.33& \colorbox[rgb]{1,1,0.4}{7.43}& 15.75&  10.27 &13.87&\colorbox[rgb]{1,1,0.4}{7.94}&10 h\\
	
	 \mvas&8.56& 9.28& 17.63& 7.00&\colorbox[rgb]{1,1,0.4}{11.86}& 13.82&12.81& 12.08&12.72&10.55&11 h\\
	
	\pisr &\colorbox[rgb]{1,0.7,0.7}{8.23}& 6.71& 16.79& \colorbox[rgb]{1,1,0.4}{5.89}&\colorbox[rgb]{1,1,0.4}{11.23}& 10.31&12.18& \colorbox[rgb]{1,1,0.4}{8.49}&\colorbox[rgb]{1,1,0.4}{12.11}&9.20& 0.5 h \\
	\gsror&12.50 &  \colorbox[rgb]{1,1,0.4}{5.30}  &\colorbox[rgb]{1,0.7,0.7}{14.68}&5.97&18.82&16.40&21.55&19.87&16.89&11.89& 1.5 h \\
	 \nero &\colorbox[rgb]{1,1,0.4}{8.41}& \colorbox[rgb]{1,0.7,0.7}{4.88}& \colorbox[rgb]{1,1,0.4}{15.29}& \colorbox[rgb]{1,0.7,0.7}{5.39}& \colorbox[rgb]{1,0.7,0.7}{9.40}& \colorbox[rgb]{1,0.7,0.7}{7.81}& \colorbox[rgb]{1,0.7,0.7}{8.34}& \colorbox[rgb]{1,0.7,0.7}{6.30}&\colorbox[rgb]{1,0.7,0.7}{10.36}&\colorbox[rgb]{1,0.7,0.7}{6.10}&8 h\\

    \midrule
    
    \dr&19.53&15.87&17.08&27.67&31.23&22.03&30.85&18.20 &24.67&20.94& 12 min \\

    \relight &16.69&14.66&17.85&13.35&29.19&15.66&26.39&20.20&22.53&15.97& 25 min \\
	
	\gs &\colorbox[rgb]{1,1,0.4}{12.79}&\colorbox[rgb]{1,1,0.4}{9.09}&\colorbox[rgb]{1,1,0.4}{16.19}&\colorbox[rgb]{1,0.7,0.7}{7.01}&\colorbox[rgb]{1,1,0.4}{24.77}&\colorbox[rgb]{1,1,0.4}{10.63}&\colorbox[rgb]{1,1,0.4}{22.92}& \colorbox[rgb]{1,1,0.4}{10.78}&\colorbox[rgb]{1,1,0.4}{19.17}&\colorbox[rgb]{1,1,0.4}{9.38}&5 min  \\
    
\polgs~(Ours) &\colorbox[rgb]{1,0.7,0.7}{10.78}& \colorbox[rgb]{1,0.7,0.7}{7.16} &\colorbox[rgb]{1,0.7,0.7}{14.63}&\colorbox[rgb]{1,1,0.4}{7.15}&\colorbox[rgb]{1,0.7,0.7}{11.51}&\colorbox[rgb]{1,0.7,0.7}{7.52}&\colorbox[rgb]{1,0.7,0.7}{12.30}&\colorbox[rgb]{1,0.7,0.7}{8.58} &\colorbox[rgb]{1,0.7,0.7}{12.31}&\colorbox[rgb]{1,0.7,0.7}{7.60}&10 min\\				
\midrule
PolGS~\citep{han2025polgs} w/o $\mathcal{L}_{pol}$ & 10.95 & 7.92 & 15.16 & 7.56 & 11.87 & 7.83 & 12.90 & 9.41 & 12.72 & 8.18 & 7 min \\
PolGS~\citep{han2025polgs} &10.88& 7.76& 15.03& 7.48& 11.55& 7.76& 12.73& 9.29& 12.55&8.07& 7 min \\
    \bottomrule
\end{tabular}	
}
\end{table*}
 \section{Experiments}
\label{sec:exp}
To evaluate the performance of our method, we conduct the following experiments: 1) Shape reconstruction
on synthetic dataset, 2) Shape reconstruction on real-world dataset, 3) Ablation study, 4) Radiance decomposition and 5) Failure cases.

\vspace{-1.4em}\paragraph{Dataset\quad}  We use four datasets to evaluate our method: the synthetic dataset \smvp, the real-world dataset \rmvp,
\pandora and \pisr, where \pandora can only be used for qualitative evaluation due to the lack of ground truth.

\vspace{-1.0em}
\paragraph{Baselines\quad} 
We compare our method with state-of-the-art techniques, including 3DGS methods \gs, \dr, and \relight, and SDF-based methods \nero, \mvas, \pandora, \nersp, \pisr, and \gsror(which surface strongly depends on SDF representation). 
All of these methods, except \gs, can handle reflective surfaces. \mvas, \pandora, \pisr, and \nersp utilize 
polarimetric information to reconstruct shapes. Specifically, we do not add the normal prior in \gs among the whole experiments for fair comparisons.
\vspace{-1em}
\begin{figure*}
	\LARGE
	\begin{overpic}
    [width=\textwidth]{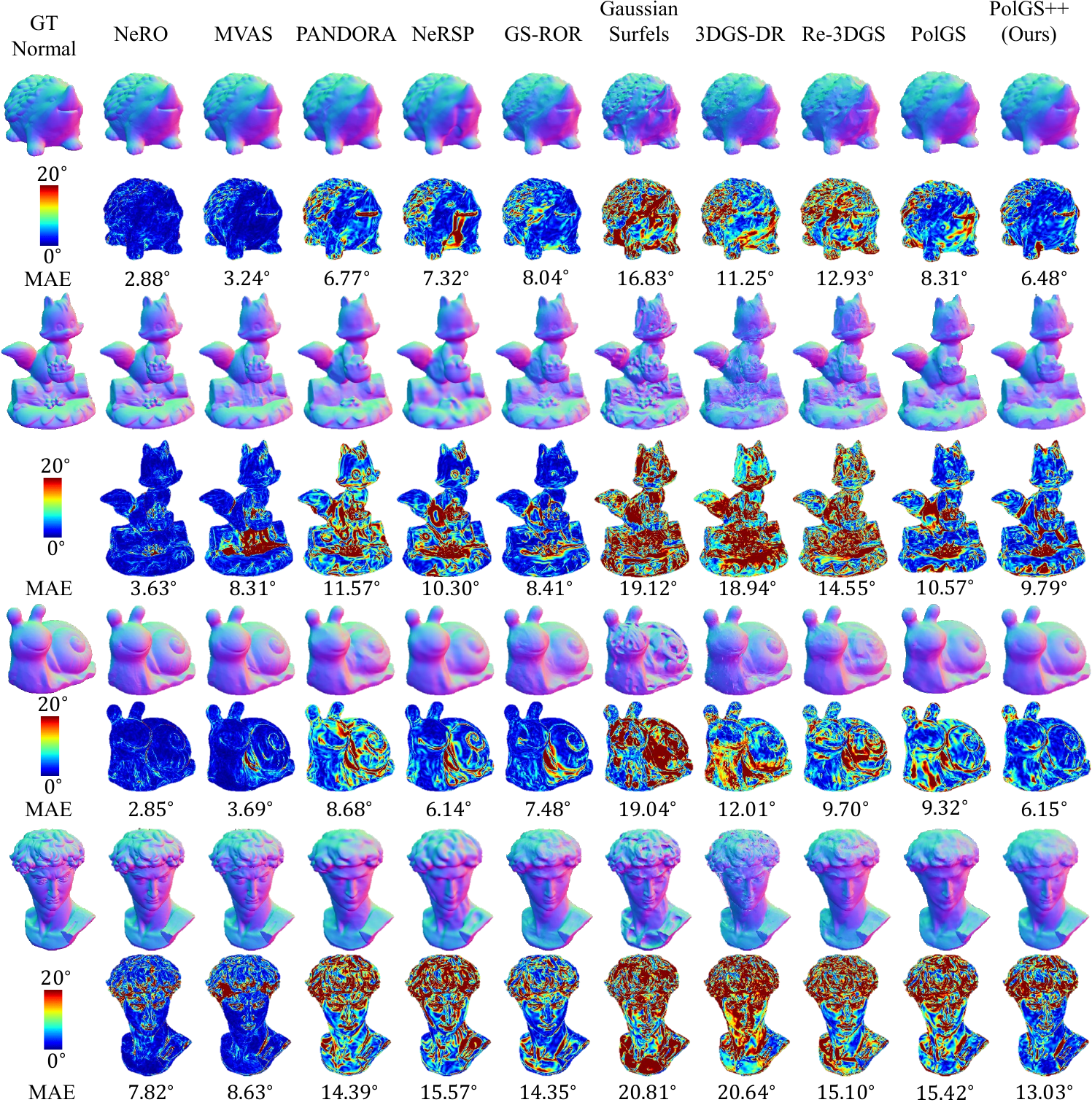}
    \put(8.8,94.5){\color{black}{\fontsize{6pt}{1pt}\selectfont {\cite{liu2023nero}}}}
    \put(18.3,94.5){\color{black}{\fontsize{6pt}{1pt}\selectfont {\cite{cao2023multi}}}}
    \put(27,94.5){\color{black}{\fontsize{6pt}{1pt}\selectfont {\cite{dave2022pandora}}}}
    \put(36.5,94.5){\color{black}{\fontsize{6pt}{1pt}\selectfont {\cite{han2024nersp}}}}
    \put(45.5,94.5){\color{black}{\fontsize{6pt}{1pt}\selectfont {\cite{zhu2025gs}}}}
    \put(54.2,94.5){\color{black}{\fontsize{6pt}{1pt}\selectfont {\cite{dai2024high}}}}
    \put(63.7,94.5){\color{black}{\fontsize{6pt}{1pt}\selectfont {\cite{ye20243d}}}}
    \put(72.2,94.5){\color{black}{\fontsize{6pt}{1pt}\selectfont {\cite{gao2023relightable}}}}
    \put(81.5,94.5){\color{black}{\fontsize{6pt}{1pt}\selectfont {\cite{han2025polgs}}}}
	\end{overpic}
	\caption{Qualitative comparisons on surface normal in \smvp, where our method can outperform existing methods based on the same representation and achieves comparable results with SDF-based methods.
	}
	\vspace{-0.5em}
	\label{fig:syn_normal}
    
\end{figure*}
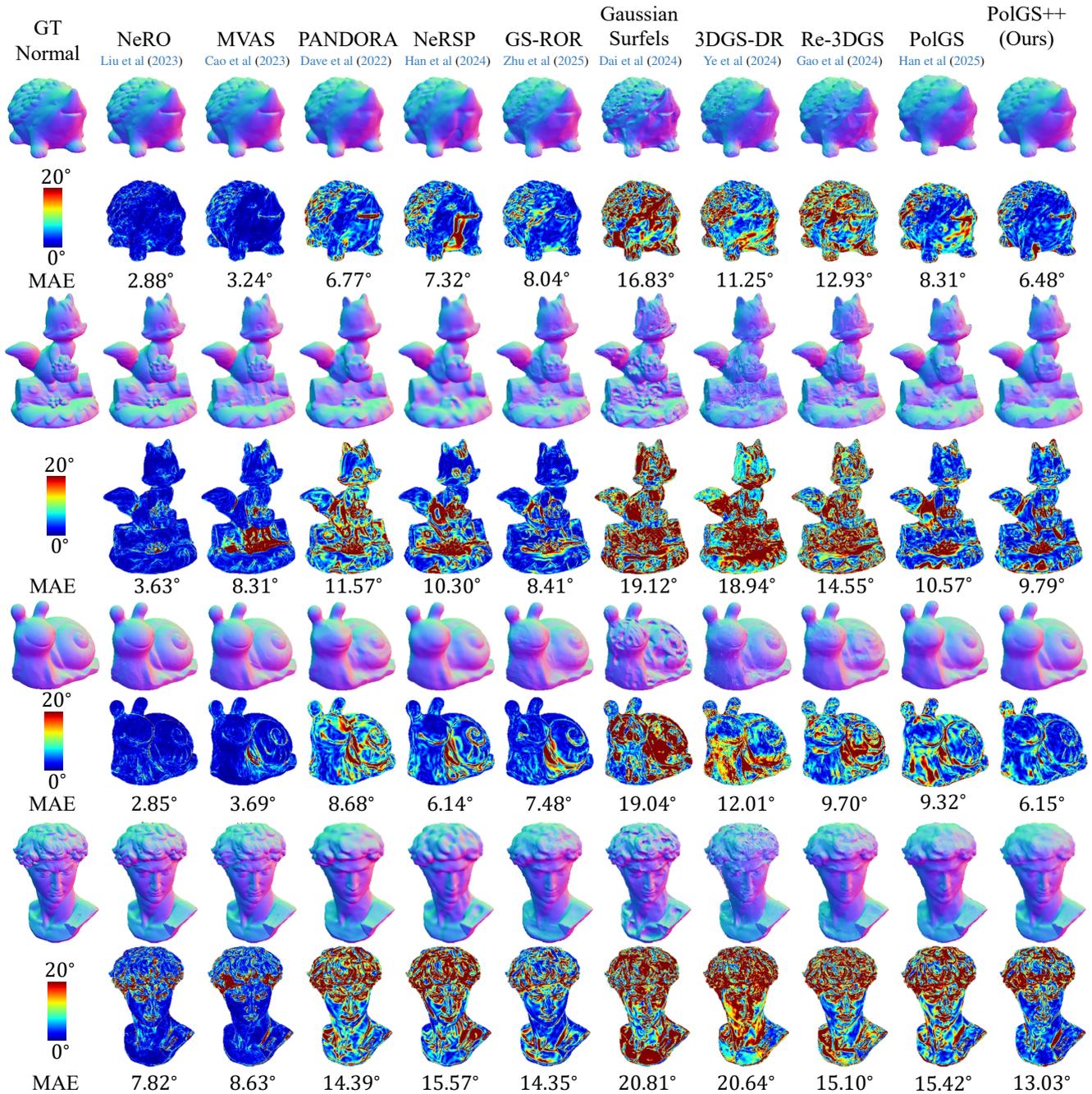
\paragraph{Evaluation metrics\quad}
We use the Chamfer Distance (CD) to evaluate the shape recovery performance and 
the mean angular error (MAE) to evaluate the quality of surface normal estimations. 3DGS-based methods provide point cloud results and we use Poisson surface reconstruction to generate the mesh for the evaluation.
\vspace{-1.2em}
\begin{figure*}
	\huge 
	\begin{overpic}
    [width=\textwidth]{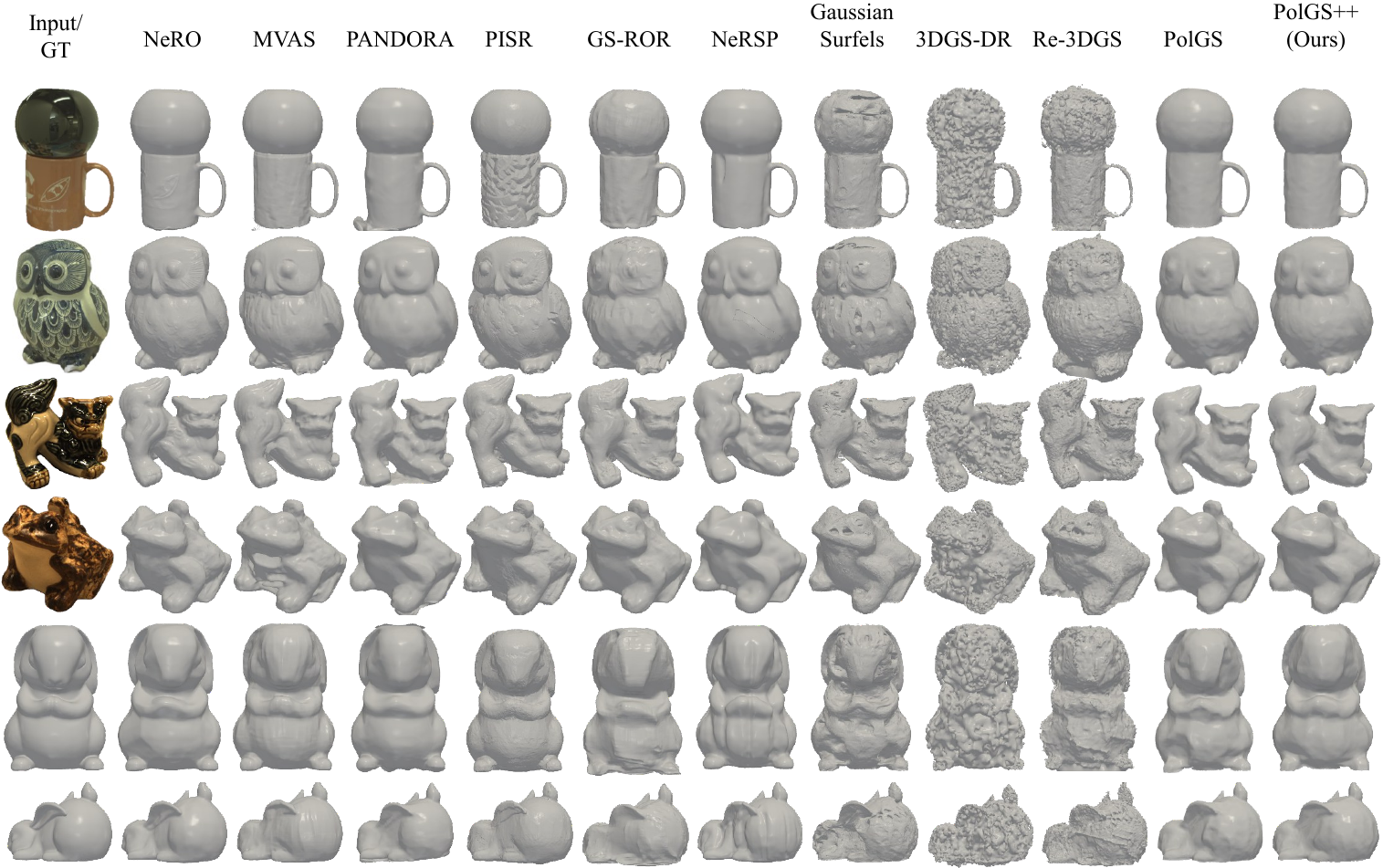}
    \put(8.8,57.8){\color{black}{\fontsize{5.5pt}{1pt}\selectfont {\cite{liu2023nero}}}}
    \put(16.5,57.8){\color{black}{\fontsize{5.5pt}{1pt}\selectfont {\cite{cao2023multi}}}}
    \put(25,57.8){\color{black}{\fontsize{5.5pt}{1pt}\selectfont {\cite{dave2022pandora}}}}
    
    \put(33.7,57.8){\color{black}{\fontsize{5.5pt}{1pt}\selectfont {\cite{chen2024pisr}}}}
    \put(42,57.8){\color{black}{\fontsize{5.5pt}{1pt}\selectfont {\cite{zhu2025gs}}}}
    \put(50,57.8){\color{black}{\fontsize{5.5pt}{1pt}\selectfont {\cite{han2024nersp}}}}
    
    \put(58,57.8){\color{black}{\fontsize{5.5pt}{1pt}\selectfont {\cite{dai2024high}}}}
    \put(66.5,57.8){\color{black}{\fontsize{5.5pt}{1pt}\selectfont {\cite{ye20243d}}}}
    \put(74.5,57.8){\color{black}{\fontsize{5.5pt}{1pt}\selectfont {\cite{gao2023relightable}}}}
    \put(83.0,57.8){\color{black}{\fontsize{5.5pt}{1pt}\selectfont {\cite{han2025polgs}}}}
	\end{overpic}
	\caption{Qualitative comparison shape on \pandora, \rmvp and \pisr, where PolGS produces similar quality of reconstruction mesh compared to SDF-based methods. }
	\label{fig:PANDORA-result}
    \vspace{-0.5em}
\end{figure*}
\paragraph{Implementation details\quad}
We conduct the experiments on an NVIDIA RTX 4090 GPU with 24GB memory. The training of our model is implemented in 
PyTorch 1.12.1~\citep{paszke2019pytorch} using the Adam optimizer~\cite{kingma2014adam}. The specific learning rates of different components
in Gaussian kernels are set the same as \gs. We adopt a warm-up strategy to train the model, and we add the polarimetric information and deferred rendering after 1000 iterations.

\subsection{Reconstruction results on synthetic dataset}

In \fref{fig:syn_normal} and \Tref{table:shape_quantitative_smvp}, we compare the shape recovery performance of various methods on the \smvp, which contains four objects with spatially varying and reflective properties. However, it is worth noting that SDF-based methods, such as \nero, \gsror, \mvas, \pandora, and \nersp, outperform 3DGS methods in terms of surface representation. This advantage is largely due to their superior ability to model complex surface details. Among the 3DGS methods, \gs struggles with reflective surfaces, while \dr and \relight still have difficulty in representing surface normals accurately.
In contrast, \polgs effectively leverages polarimetric information, significantly improving geometric surface performance. Due to the inadequate point cloud generation by other 3DGS methods, they fail to produce reasonable mesh results using Poisson surface reconstruction. Our method, however, achieves the lowest mean Chamfer Distance across the synthetic \smvp dataset, reinforcing the trends observed in surface normal estimations and confirming that our approach delivers the best performance among the 3DGS techniques.

\subsection{Reconstruction results on real-world dataset}
We further evaluate the reconstruction performance on the \pandora, \rmvp,  and \pisr datasets, with qualitative results illustrated in \fref{fig:PANDORA-result} and quantitative results detailed in \Tref{table:shape_quantitative_rmvp_pisr}. In real-world scenarios, our method produces reconstructions that align more closely with SDF-based approaches, while significantly outperforming other 3DGS-based methods in terms of reconstruction quality. For example, in the {\sc Vase} case, our approach accurately estimates the shape of a ceramic surface in just $10$ minutes, achieving results closer to those generated by SDF-based methods. 
Additionally, the {\sc Frog} sample highlights our method's ability to reconstruct objects with rough glossy surfaces, showing the robustness and generalization capability of our method. 
Furthermore, the texture-less objects in the \pisr dataset also demonstrate the advantage of our method in handling objects with limited RGB cues, where the polarimetric information provides critical constraints for accurate shape recovery.
These results collectively demonstrate the effectiveness of our approach in handling diverse real-world objects with varying surface properties.

\begin{figure*}
	\centering
	\includegraphics[width=\textwidth]{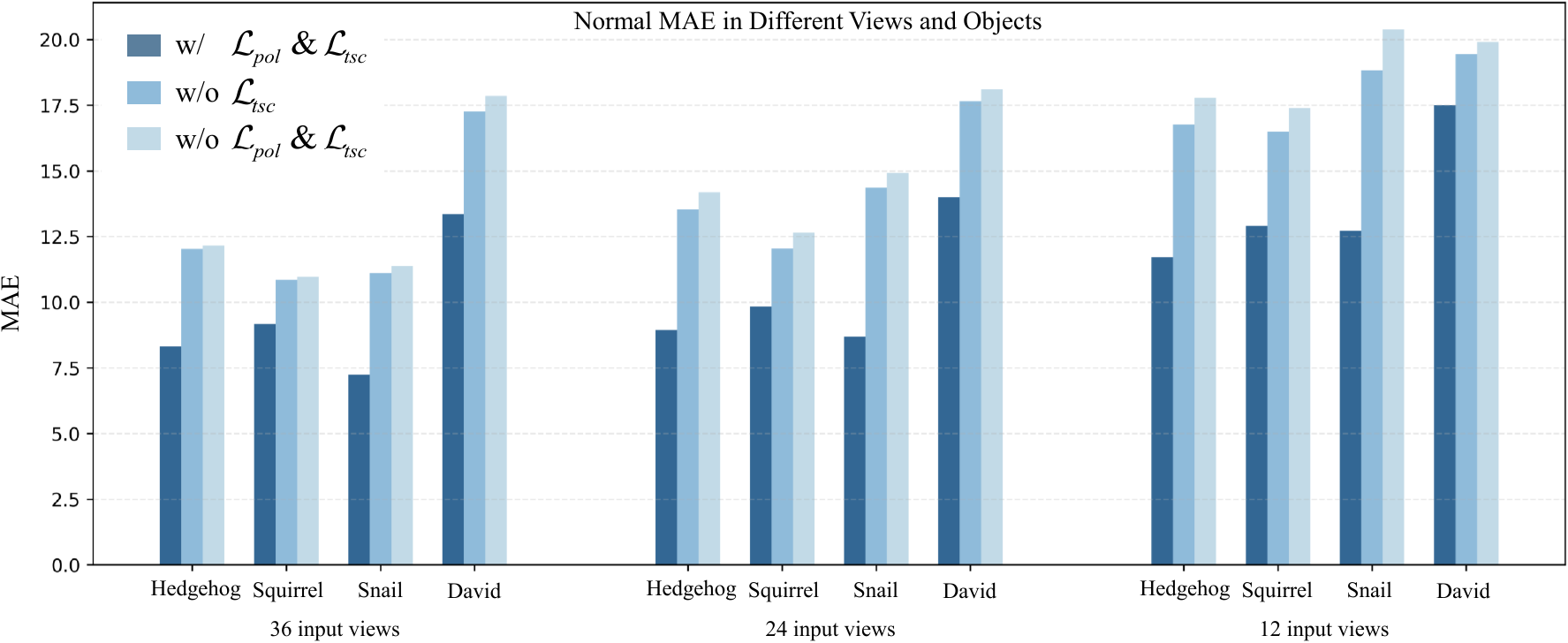}
	\caption{ Quantitative ablation results on \smvp, where we evaluate the effect of polarimetric supervision under different input views. }

	\label{fig:ablation_table}
	
\end{figure*}
\begin{figure}
	\centering
	\includegraphics[width=\linewidth]{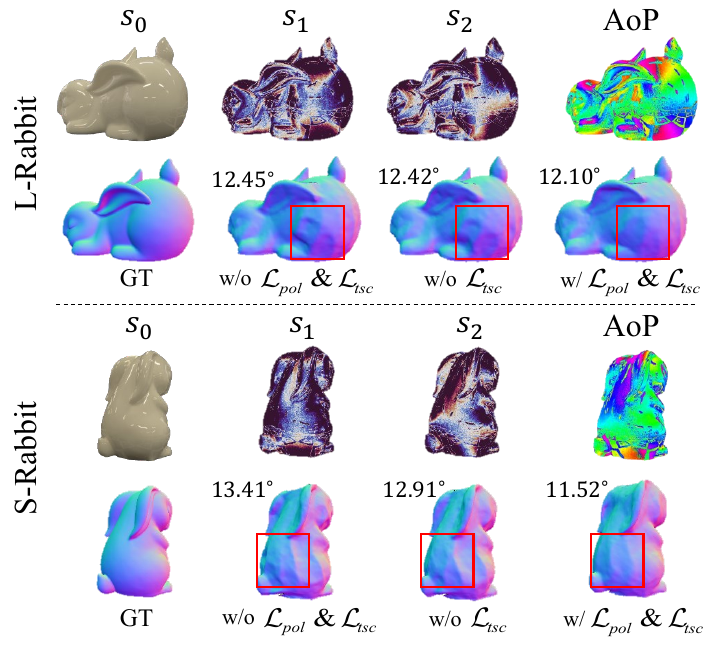}
	\caption{
Qualitative ablation results on \pisr, where we compare the reconstruction quality with and without polarimetric supervision. }

	\label{fig:ablation}
	\vspace{-1em}
\end{figure}
\subsection{Ablation study}\label{sec:ablation}

In this section, we conduct three ablation studies to validate key design choices in our method.

\begin{table}[t]
\centering
\caption{Contribution of each module, progressing from the Gaussian Surfels base to the full \polgs, on \smvp.}
\label{table:module_ablation}
\begin{tabular}{lcc}
\toprule
Variant & MAE$\downarrow$ & CD$\downarrow$ \\
\midrule
(a) Gaussian Surfels & 19.69 & 9.95 \\
(b) PolGS w/o $\mathcal{L}_{pol}$ & 12.12 & 7.75 \\
(c) PolGS & 11.47 & 7.51 \\
(d) \polgs, w/o visibility & 10.88 & 7.19 \\
(e) \polgs, SDF-oracle visibility & 9.42 & 6.64 \\
(f) \polgs, depth-guided visibility & 9.53 & 6.85 \\
\bottomrule
\end{tabular}
\vspace{-0.5em}
\end{table}

\paragraph{Component-wise ablation\quad}
\Tref{table:module_ablation} quantifies the contribution of each component along the development from Gaussian Surfels to \polgs. Introducing the pBRDF-based PolGS representation reduces MAE/CD from $19.69/9.95$ to $12.12/7.75$, while adding $\mathcal{L}_{pol}$ further improves the result to $11.47/7.51$. On top of PolGS, TSC without visibility gating reaches $10.88/7.19$, and the depth-guided visibility mask further reduces the errors to $9.53/6.85$. This result is close to the SDF-oracle visibility reference of $9.42/6.64$, showing that the proposed depth-guided mask provides reliable visibility reasoning. These results separately verify the effectiveness of the pBRDF-based representation, $\mathcal{L}_{pol}$, TSC, and depth-guided visibility.

\paragraph{Effect of polarimetric information\quad}
We first study how polarimetric cues contribute to surface reconstruction. Our polarimetric supervision consists of two terms: a photometric polarization loss $\mathcal{L}_{pol}$ and a tangent-space consistency loss $\mathcal{L}_{tsc}$. We perform controlled experiments on \smvp by varying both the number of input views and the enabled loss terms (w/o polarimetric loss, w/ $\mathcal{L}_{pol}$ only, and w/ both).

\Fref{fig:ablation_table} shows a clear trend: when view coverage is dense, RGB photometric cues already provide strong constraints, and the additional gain from $\mathcal{L}_{pol}$ becomes marginal. In contrast, $\mathcal{L}_{tsc}$ yields a more consistent improvement, indicating that enforcing stable polarization-derived constraints is particularly beneficial even in well-observed settings. As the number of views decreases, the advantage of polarimetric information becomes increasingly evident, and combining $\mathcal{L}_{pol}$ with $\mathcal{L}_{tsc}$ leads to the most robust reconstructions under sparse-view conditions.
\begin{table*}[t]
	\centering
	\caption{Sensitivity to the visibility-mask threshold $\tau$ on all datasets with ground truth. Each entry reports MAE/CD; the same $\tau$ is used for every dataset.}
	\label{table:visibility_ablation}
	\resizebox{0.96\linewidth}{!}{%
		\begin{tabular}{lccccccc}
			\toprule
			Dataset & 0.002 & 0.005 & 0.008 & \textbf{0.010} & 0.015 & 0.020 & w/o mask \\
			\midrule
			SMVP3D & 9.64/6.93 & 9.60/6.91 & 9.55/6.88 & \textbf{9.53/6.85} & 9.54/6.87 & 9.59/6.89 & 10.88/7.19 \\
			RMVP3D & 12.82/7.25 & 12.78/7.23 & 12.74/7.19 & \textbf{12.71/7.16} & 12.73/7.19 & 12.76/7.22 & 13.48/7.41 \\
				PISR & 12.12/8.18 & 11.98/8.15 & 11.94/8.13 & \textbf{11.91/8.05} & 11.93/8.12 & 11.97/8.15 & 12.69/8.36 \\
			\bottomrule
	\end{tabular}}
\end{table*}
We further provide a qualitative evaluation on \pisr to highlight the practical value of polarization in \fref{fig:ablation}. While the RGB appearance lacks discernible texture, the polarization channels $s_1$ and $s_2$ and AoP information exhibit meaningful variations that provide additional geometric cues for resolving shape ambiguities. The results show that both concave and convex surface distortions are more accurately reconstructed when polarization information is employed.

\begin{figure}
    \centering
    \includegraphics[width=\linewidth]{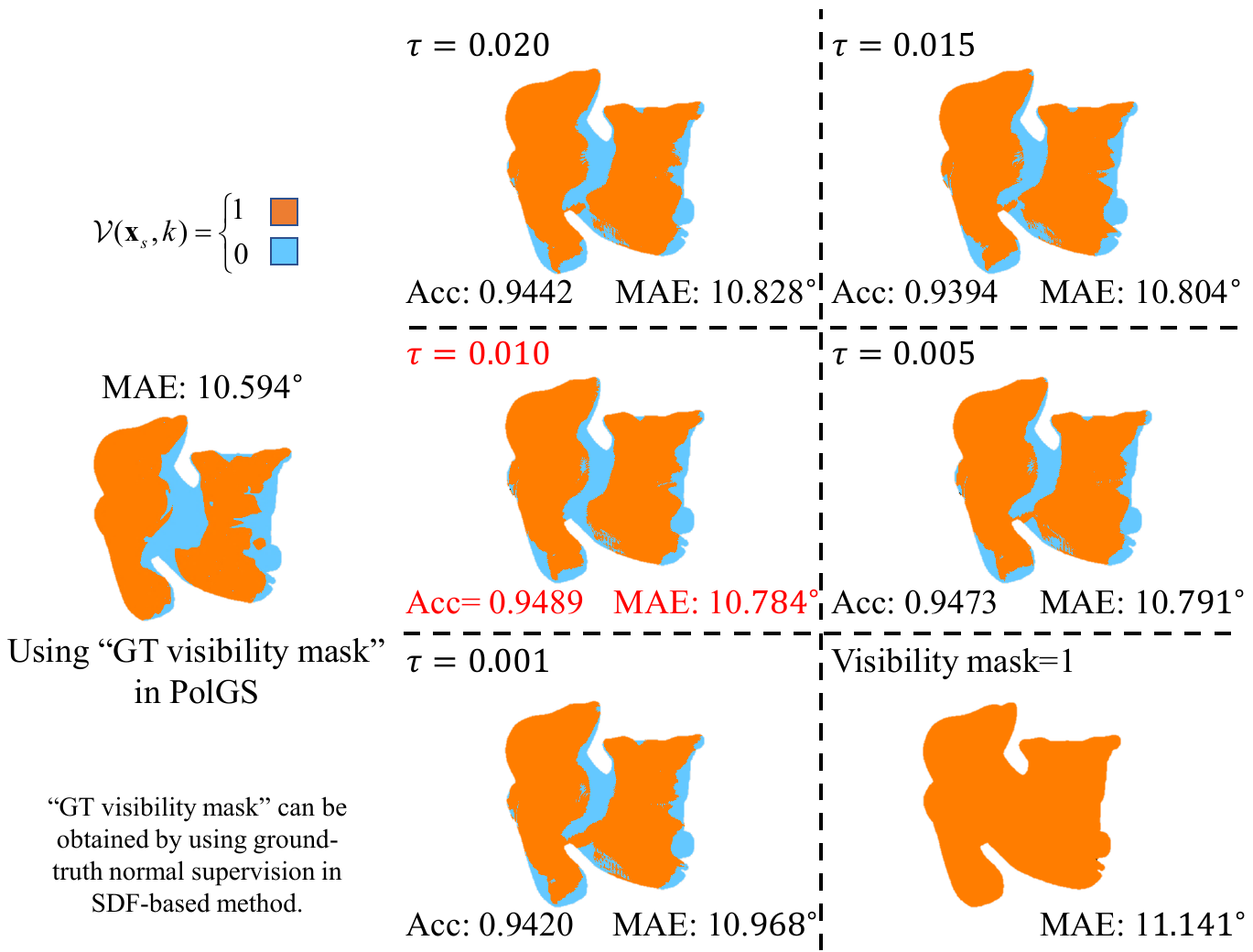}
    \caption{Ablation results on the visibility mask threshold.}
    \label{fig:visibility_ablation}
    \vspace{0em}
\end{figure}

\paragraph{Visibility mask threshold\quad}
\label{exp:visibility}
We then validate the threshold $\tau$ used to binarize the depth-guided visibility mask in \Eref{eq:visibility}. We sweep different thresholds and compare the resulting masks to the SDF-based visibility mask under ground-truth normal supervision.

As shown in \fref{fig:visibility_ablation}, directly using the SDF-based visibility masks yields a surface normal MAE of 10.594 degrees in \polgs. With depth-guided masks, performance varies with $\tau$, and $\tau = 0.010$ provides the best trade-off: an accuracy of 0.9489 against the GT visibility mask and an MAE of 10.784 degrees. Without any visibility mask, the MAE degrades to 11.141 degrees, confirming the benefit of incorporating visibility information.

We further evaluate the sensitivity of the final reconstruction to $\tau$ on all datasets with ground truth. As reported in \Tref{table:visibility_ablation}, the same setting $\tau=0.010$ achieves the best MAE/CD on SMVP3D, RMVP3D, and PISR. The results vary only slightly over $\tau\in[0.002,0.020]$, whereas removing the visibility mask consistently causes a clearer degradation. Together, the mask-level analysis in \fref{fig:visibility_ablation} and the cross-dataset reconstruction results in \Tref{table:visibility_ablation} show that the visibility gate is beneficial and that its performance is not sensitive to the precise threshold. We therefore use $\tau=0.010$ for all experiments without dataset-specific tuning.

\vspace{-1em}

\subsection{Comparison of radiance decomposition}
\begin{figure}
	\begin{overpic}[width=\linewidth, trim={0pt 0pt 0pt 0pt}]{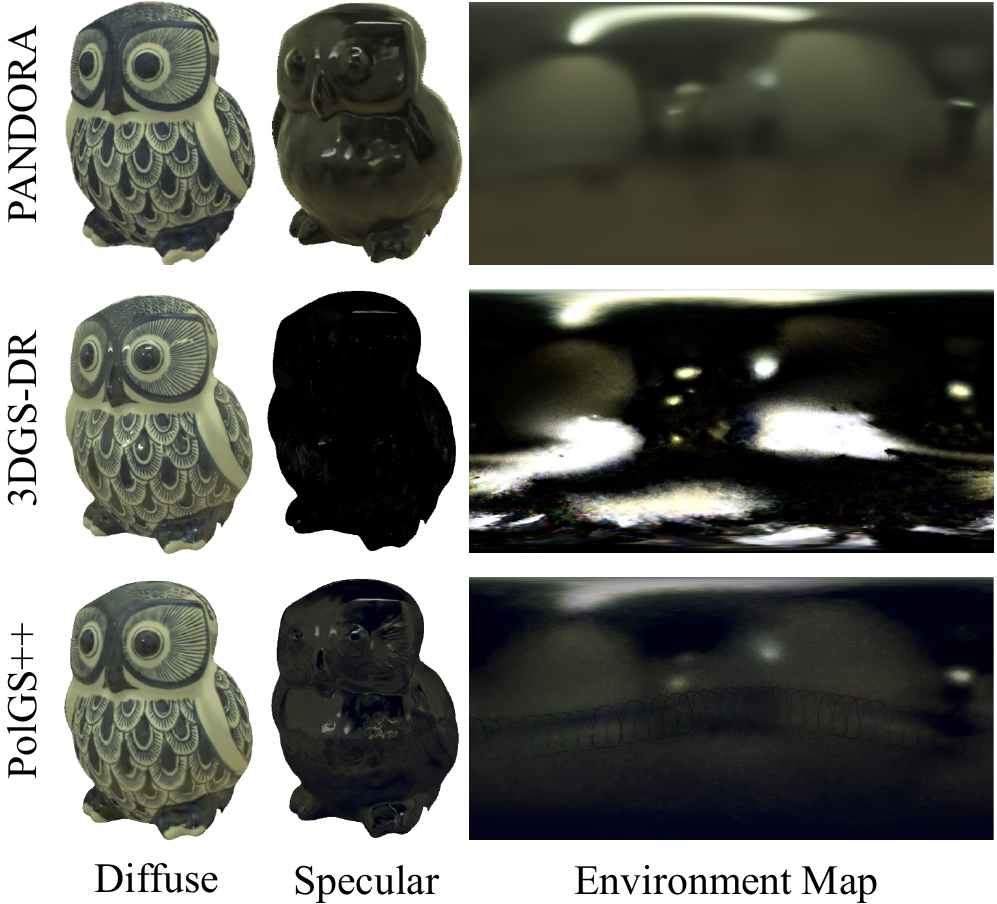}
		
	\end{overpic}
	\vspace{-1.5em}
	\caption{Separation of diffuse and specular components with \pandora, \dr and \polgs.}
	\label{fig:env_compare}
	\vspace{-0em}
\end{figure}

\Fref{fig:env_compare} presents a comparison of diffuse and specular component decomposition generated by \pandora, \dr, and \polgs. Here, \pandora utilizes the IDE~\cite{verbin2022ref} structure to produce environment map results, whereas \polgs adopts \dr's methods by employing a Cubemap encoder for the same purpose. 
Compared with \dr, \polgs leverages additional polarimetric information to effectively constrain and disambiguate diffuse and specular components. 
Notably, our results closely align with \pandora's, demonstrating improved radiance decomposition ability.

\begin{figure}[t]
	\centering
	\includegraphics[width=\linewidth]{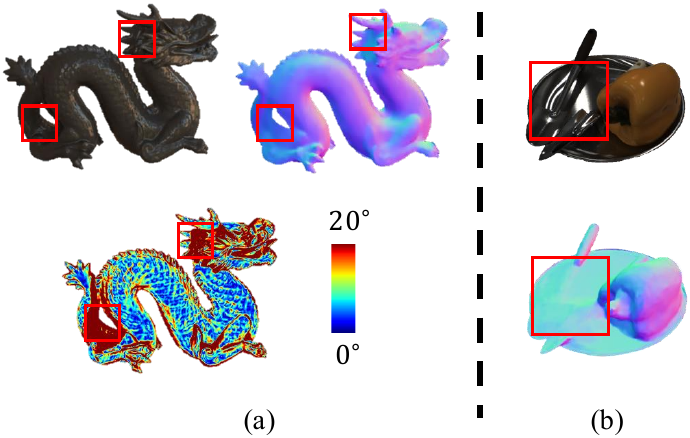}
	\caption{Representative failure cases. (a) A complex object with significant inter-reflection: the input image and estimated normal are shown in the top row, and the angular-error map ($0^\circ$--$20^\circ$) is shown below. The red boxes highlight concave and intricate regions with large reconstruction errors. (b) A metallic object: the red box highlights the region where the estimated normal is distorted because of the mismatch between conductor reflection and the dielectric pBRDF used in our current model.}
	\label{fig:failure}
\end{figure}
\vspace{-1em}
\subsection{Failure cases}
\label{sec:failure}
We present two representative failure cases in \Fref{fig:failure}. First, the current model primarily describes local, approximately single-bounce reflection under fixed environment lighting. For complex surfaces with significant inter-reflection or refraction, the Stokes observation at a pixel can mix multiple light paths and is no longer determined by a single local normal. This violates the local formation assumptions used by both pBRDF rendering and AoP/TSC, causing local surface distortion or loss of detail. As shown in \Fref{fig:failure}(a), the largest angular errors concentrate around the concave and intricate regions highlighted by the red boxes.

Second, the current pBRDF primarily targets dielectric materials. Conductors generally have a complex refractive index, and the resulting material-model mismatch biases appearance decomposition and normal optimization. In \Fref{fig:failure}(b), the reconstructed normal is visibly distorted around the metallic component because the dielectric pBRDF does not correctly describe conductor reflection. Addressing these cases requires a more general conductor pBRDF and a model of multi-bounce polarized light transport.

Both cases are non-ideal conditions under which the assumptions of the proposed pBRDF model no longer hold, and the measured AoP no longer corresponds to a single local surface normal. The same degradation occurs when the degree of polarization is low, such as under weak illumination, or when the environment lighting varies during capture.
 \section{Conclusion}
\label{sec:conclusion}
We propose \polgs, a physically-guided polarimetric Gaussian Splatting method for reflective surface reconstruction. We leverage polarimetric cues to separate diffuse and specular components from the Stokes field, providing physically grounded constraints for surface normal estimation in the Gaussian Splatting representation. Furthermore, we introduce a depth-guided visibility mask acquisition method to enforce tangent-space consistency with AoP information, leading to improved reconstruction quality on reflective surfaces.
Extensive experiments demonstrate that \polgs outperforms state-of-the-art 3DGS-based methods in reconstruction accuracy while retaining high efficiency.

\paragraph{Limitations\quad} While \polgs achieves strong results on reflective surfaces, it relies on accurate polarimetric measurements, which can be affected by noise and calibration errors. 
Moreover, \polgs currently assumes a fixed environmental lighting condition and cannot handle dynamic lighting scenarios. 
Additionally, our method mainly focuses on dielectric materials.
\Sref{sec:failure} presents two representative failure cases, a complex object with significant inter-reflection and a metallic object, that illustrate these limitations.
Future work could explore more robust polarimetric estimation techniques and extend the model to handle a wider range of materials and lighting conditions.

\section*{Statements and Declarations}

\textbf{Funding}

\noindent{This work was supported by Beijing Municipal Science \& Technology Program No. Z251100007125021, National Natural Science Foundation of China (Grant No. 62472044, U24B20155, U23B2052).Hebei Natural Science Foundation Project No. 242Q0101Z, Beijing-Tianjin-Hebei Basic Research Funding Program No. F2024502017, Beijing Natural
Science Foundation Project No. L242025, BUPT Excellent Ph.D. Students Foundation.}
\vspace{1em}

\noindent{\textbf{Competing Interests}}

\noindent The authors have no relevant financial or non-financial interests to disclose.

\vspace{1em}
\noindent \textbf{Data Availability}

\noindent The source code and datasets used in this study will be made publicly available upon formal acceptance of the manuscript.

\newpage
 
{\small
\bibliographystyle{spbasic}
\bibliography{egbib}
}

\end{document}